\documentclass{article}

\usepackage[final,nonatbib]{neurips_2024}

\usepackage[utf8]{inputenc} 
\usepackage[T1]{fontenc}    
\usepackage{hyperref}       
\usepackage{url}            
\usepackage{booktabs}       
\usepackage{amsfonts}       
\usepackage{pifont}         
\usepackage{nicefrac}       
\usepackage{microtype}      
\usepackage{xcolor}         
\usepackage{amsmath}
\usepackage{amsthm}
\usepackage{wrapfig}
\usepackage[english]{babel}
\usepackage[square,numbers]{natbib}
\usepackage{geometry}
\usepackage{booktabs}
\usepackage{array}
\usepackage{longtable}
\usepackage{soul}
\usepackage{graphicx}
\usepackage{amsmath}
\usepackage{caption}

\newcommand{\cmark}{\ding{51}}  
\newcommand{\xmark}{\ding{55}}  

\usepackage{amsthm}
\usepackage{mdframed}

\newmdtheoremenv[
  linecolor=black,
  linewidth=1pt,
  backgroundcolor=black!5,
  topline=true,
  bottomline=true,
  leftline=true,
  rightline=true
]{theorem}{Proposition}
\newmdtheoremenv[
  linecolor=black,
  linewidth=1pt,
  backgroundcolor=black!5,
  topline=true,
  bottomline=true,
  leftline=true,
  rightline=true
]{theorem_proof}{Proposition}
\newmdtheoremenv[
  linecolor=gray,
  linewidth=1pt,
  backgroundcolor=gray!5,
  topline=true,
  bottomline=true,
  leftline=true,
  rightline=true
]{basic_proof}{Property}
\newmdtheoremenv[
  linecolor=blue,
  linewidth=1pt,
  backgroundcolor=blue!5,
  topline=true,
  bottomline=true,
  leftline=true,
  rightline=true
]{lemma}{Lemma}
\title{A Circular Argument:\\Does RoPE \textit{need} to be Equivariant for Vision?}

\author{%
Chase van de Geijn$^{1,\dagger}$,\quad Timo L\"uddecke$^{1}$,\quad Polina Turishcheva$^1$ \quad Alexander S. Ecker$^{1,2,*}$ \\
$^1$Institute of Computer Science and Campus Institute Data Science, University of Göttingen \\
$^2$Max Planck Institute for Dynamics and Self-Organization, Göttingen, Germany\\
$^\dagger$\texttt{chase.geijn@uni-goettingen.de}\\
$^*$\texttt{ecker@cs.uni-goettingen.de}
}

\begin{document}

\maketitle

\begin{abstract}
    Rotary Positional Encodings (RoPE) have emerged as a highly effective technique for one-dimensional sequences in Natural Language Processing spurring recent progress towards generalizing RoPE to higher-dimensional data such as images and videos. 
    The success of RoPE has been thought to be due to its positional equivariance, i.e. its status as a \textit{relative} positional encoding. In this paper, we mathematically show RoPE to be one of the most general solutions for equivariant positional embedding in one-dimensional data. Moreover, we show Mixed RoPE to be the analogously general solution for $M$-dimensional data, if we require commutative generators -- a property necessary for RoPE's equivariance. However, we question whether strict equivariance plays a large role in RoPE's performance. We propose Spherical RoPE, a method analogous to Mixed RoPE, but assumes non-commutative generators. Empirically, we find Spherical RoPE to have the equivalent or better learning behavior compared to its equivariant analogues. This suggests that relative positional embeddings are not as important as is commonly believed, at least within computer vision. We expect this discovery to facilitate future work in positional encodings for vision that can be faster and generalize better by removing the preconception that they must be relative.
\end{abstract}

\section{Introduction}

Deep learning is in the age of transformers \cite{vaswani2017attention}. At their core, transformers are built on attention \cite{bahdanau2014neural, schmidhuber1992learning}, which is a permutation-invariant operation \cite{velivckovic2017graph}, making them agnostic to word or token position within a corpus. To break this symmetry, tokens must be modified with position embeddings 
\cite{he2023sheaf,you2019position}. Recently, Rotational Positional Encodings (RoPE) \cite{su2024roformer} have gained popularity, touting an emphasis on the \textit{relative} position between two tokens rather than their absolute positions \cite{grattafiori2024llama,heo2024rotary,liu2024deepseek,liu2023molrope}. However, some of the original claims of RoPE have been called into question leading to confusion as to \textit{why} it works: \citet{su2024roformer} claimed the attention scores to decay with distance between tokens. This was found to be true only for attention with the same query and key \cite{barbero2024round}. Moreover, transformers with causal masking have been shown to require no positional encodings to be capable of recovering absolute position \cite{haviv2022transformer}, making RoPE's relative (shift-equivariant) claim questionable. However, many new methods continue to be motivated by RoPE’s benefit from shift-equivariance \cite{islam2025platonic,schenck2025learning, yu2025comropescalablerobustrotary} . To guide future research in positional encodings, it is important to discover whether shift-equivariance truly makes RoPE successful and needs to be preserved when extending it.

Both transformer and RoPE were originally designed for one-dimensional sequences such as language. RoPE encodes position by pairing dimensions within the query and key vectors within a transformer and rotating the paired dimensions. Transformers have become the current staple across all AI fields \cite{dosovitskiy2020image,jumper2021highly,liu2023molrope, liu2021swin,shamshad2023transformers}.
Naturally, RoPE's recent popularity in NLP has also spread to Vision Transformers (ViT), where the data is two- or three-dimensional, corresponding to images and videos. 
\textit{How} to extend RoPE to other modalities is nontrivial and assumptions must be made to maintain equivariance \cite{liu2025rethinking,schenck2025learning}. 
The most commonly used approach for extending RoPE to ViTs is through Axial  RoPE \cite{chu2024visionllama, heo2024rotary,  wei2025videorope}, partitioning the embedding dimensions into dimensions rotated independently either by the horizontal or the vertical position of the tokens. 
However, this approach does not allow for diagonal attention patterns where horizontal and vertical information ``mix", which have been hypothesized to enhance generalization; consequently, learned Mixed RoPE was proposed \cite{heo2024rotary}. 
Even more recently, LieRE~\cite{ostmeier2024liere} generalized Mixed RoPE from pair-wise rotations to higher dimensional rotations using learned skew-symmetric Lie algebras. If one defines rotations to be special orthogonal transformation, LieRE is the most general form of rotation encoding. However, while general, LieRE does not guarantee equivariance. 

In this paper, we investigate the relationships across these different forms of positional encoding. 
In Section \ref{sec:generality}, we mathematically show RoPE with parameterized rotation speeds to be equivalent to LieRE for one-dimensional data.
When the number of positional dimensions is higher dimensional, LieRE is not guaranteed to be equivariant unless constraints are placed on the Lie algebras. 
Using this insight, we derive Axial RoPE by imposing a ``mutual exclusivity" constraint on the eigenvalues of LieRE's generators. 
Further, we will show that if one loosens this constraint -- requiring the Lie algebra to be commutative between the generator -- then one arrives at Mixed RoPE.
To be a relative encoding, this commutativity property is necessary \cite{liu2025rethinking, schenck2025learning}, thus making Mixed RoPE the most general form of LieRE which maintains equivariance.
However, it has been noted that requiring the positional embedding to be relative is an inductive bias whose necessity to RoPE's success is unclear \cite{abramson2024accurate, barbero2024round,haviv2022transformer}. 

The perceived necessity of equivariance has led to a circular argument where positional embeddings are assumed to perform well because they are relative, and all new embeddings must be relative because relative embeddings perform well. To break this cycle, we believe that it is imperative to establish the importance of equivariance embeddings for multi-dimensional RoPE.
In Section \ref{sec:experiments}, we propose alternative methods to establish a cause-effect experiment to evaluate whether equivariances is a predominant contributor to RoPE's faster training dynamics and generalization. To this end, we propose Spherical RoPE which takes a non-commutative assumption, thus breaking equivariance, and Uniform RoPE, which maintains equivariance, but has only a single shared rotation speed.

In Section \ref{sec:results}, we find that Spherical RoPE has the same training behaviors as its equivariant analogues and we find that Uniform RoPE outperforms the standard learned encodings, while performing worse than other RoPE methods. We conclude that our evidence suggests that the performance of RoPE over traditional embeddings is not explained by equivariance.


\section{Background}

In this section, we review concepts and notation from previous work on rotary positional embeddings. We introduce the methods in both historical and progressively general order which we will use to prove in Section~\ref{sec:generality} that Mixed RoPE is the most general $M$-D rotary embedding with equivariance. For a broader literature review on positional embeddings see Appendix~\ref{app:litrev}. For a compact overview of symbols, see Appendix~\ref{app:notations}.

\subsection{Attention}

We use the standard attention mechanism from \citet{vaswani2017attention}, given by
\begin{equation}
    \mathbf{Z} = \text{Attention}(\mathbf{Q}, \mathbf{K}, \mathbf{V}) = \text{softmax}\left( \frac{\mathbf{Q}\mathbf{K}^\top}{\sqrt{d_k}} \right)\mathbf{V}.
\end{equation}
We consider only single-headed attention to simplify notation, so here $\mathbf{Q},\mathbf{K}$, and $\mathbf{V}$ are elements of $\mathbb{R}^{T\times N}$, where $T$ is the number of tokens and $N$ is the network's latent dimension. We will primarily use index notation, where the above equation is expressed as, $
    \mathbf{z}_i = \sum_{j=1}^\top a(\mathbf{q}_i,\mathbf{k}_j) \mathbf{v}_j.$
We define the attention mechanism, $a(\mathbf{q}_i,\mathbf{k}_j)$, as
\begin{equation}
    a(\mathbf{q}_i,\mathbf{k}_j) = \frac{e^{\alpha(\mathbf{q}_i, \mathbf{k}_j)}}{\sum_{j=1}^T e^{\alpha(\mathbf{q}_i, \mathbf{k}_j')}},
\end{equation}
where what we refer to as the \textit{attention score} is given by
\begin{equation} \label{eq:att_score}
\alpha(\mathbf{q},\mathbf{k}) = \mathbf{q}^{\top}\mathbf{k}.
\end{equation}
This formulation of attention is equivariant to permutations of the token order. To break this symmetry, the position of the tokens must be ``encoded" into the attention scores.
Thus, we re-express the attention score as a function of the content of the query token $\mathbf{x}_i\in \mathbb{R}^N$ and key token $\mathbf{x}_j\in \mathbb{R}^N$, and their positions $p_i, p_j\in \mathbb{R}$,
\begin{equation}
    \alpha_{ij} := \alpha(\mathbf{q}_i,\mathbf{k}_j) := \alpha((\mathbf{x}_i, p_i),(\mathbf{x}_j,p_j)) :=   \alpha(\mathbf{x}_i,\mathbf{x}_j, p_i,p_j).
\end{equation}
Throughout this paper, we will abuse the notation of $\alpha$ and use these expressions interchangeably for ease of notation. If the position affects the query and key directly, as in RoPE, we will introduce the notation $\alpha(\varphi(x_i,p_i), \varphi(x_j,p_j))$ for positional encoding function $\varphi$.

\subsection{Absolute and Relative Positional Encoding}


Absolute Positional Encoding (APE) is a common way of embedding token positions in transformers by adding position-dependent vectors, i.e. $\varphi(\mathbf{x}, p) := \mathbf{x} + \mathrm{PE}(p)$, where $\mathbf{x}$ is a token embedding, $p$ is its position, and $\mathrm{PE} : \mathbb{Z} \to \mathbb{R}^N$.
Previous work has suggested learning a per-position token as $\mathrm{PE}$ \cite{dosovitskiy2020image, DBLP:journals/corr/GehringAGYD17}. However, this restricts the network to fixed context length, removing the ability to extrapolate to different sequence lengths.
The alternative is to add a deterministic function to the embedding. 
\citet{vaswani2017attention} proposed to add Fourier modes,
\begin{equation}
    PE_n(p) = 
    \begin{cases}
        \sin\left(p \, \omega_{\frac{n}{2}}\right), & \text{if } n \bmod 2 = 0 \\
        \cos\left(p \, \omega_{\lfloor \frac{n}{2} \rfloor}\right), & \text{if } n \bmod 2 = 1,
    \end{cases}
    \label{eq:positional_encoding}
\end{equation}
where $n$ is a dimension within the positional embedding vector and $\omega_n$ is a frequency term which increases with dimension. 
Note that this pairs elements in the embedding vector with each pair being transformed by the same frequency. 

For ease in future notation, we will use $D:=N/2$ as the number of pairs and interpret the embedded token as a $D \times 2$ tensor. 
One can also interpret this tensor as representing the coefficients of a complex number, the first representing the real and second representing the complex part. Then we can succinctly write this form of positional encoding as
\begin{align}
    \varphi(\bar{\mathbf{x}}, p) = \bar{\mathbf{x}} + e^{i\pmb{\omega}p},
\end{align}
where we use $~\bar{\cdot}~$ to indicate complex-valued vectors, $\bar{\mathbf{x}} \in \mathbb{C}^D$.
For this notation, we should also adjust the attention score for complex numbers,
\begin{align}\label{eq:complex_form}
    \alpha(\bar{\mathbf{q}}, \bar{\mathbf{k}}) = \text{Re}\left[\bar{\mathbf{q}}^\top\bar{\mathbf{k}}\right],
\end{align}
where $\bar{\mathbf{q}}= \bar{\mathbf{W}}_q\varphi(\bar{\mathbf{x}},p)$ and $\bar{\mathbf{k}}= \bar{\mathbf{W}}_k\varphi(\bar{\mathbf{x}},p)$, with $\bar{\mathbf{W}}_q, \bar{\mathbf{W}}_k \in \mathbb{C}^{D\times D}$, and $\top$ is assumed to be the Hermitian transpose. With Eq.~\ref{eq:complex_form} implied, we will continue with the notation in Eq. \ref{eq:att_score}.

\paragraph{Relative Positional Encodings}

Positional embeddings rely on being able to assign position values to each token. However, how one assigns positions can often be arbitrary. One could just as correctly assign the first token the value zero and consider natural numbers, or assign the middle token of a corpus zero and consider integers. We can relax the assumption of a canonical way of labeling positions in APE by relying on relative distances between tokens, resulting in $\alpha_{ij} = \alpha(\mathbf{x}_i,\mathbf{x}_j, p_i - p_j)$.
This is called relative positional encoding. 
We refer to this property as embeddings having a \textit{relative positional bias}, or equivalently, having \textit{shift-equivariance} (see Appendix~\ref{app:equiv} for discussion of the equivalence). In this manuscript, we will simply use the term equivariance with the implication that the attention score is invariant to shifts in the query and key.

\subsection{Rotary Positional Encodings (RoPE)}

There are four common properties that are often preferred for positional embeddings: equivariance, key-query separability, linearity, and locality. 
For further details and why one may want these properties see Appendix~\ref{app:properties}.

From the properties, Rotary Positional Embeddings (RoPE) were derived by \citet{su2024roformer}. Rather than adding a positional embedding to the patch embedding, RoPE proposed to \textit{modify the queries and keys} by rotating them in pairs. By interpreting queries and keys as complex vectors, we can express this rotation as
\begin{align}\label{eq:rope_index}
    \varphi(\bar{q}_d, p) = e^{i\omega_dp} \bar{q}_d& & \varphi(\bar{k}_d, p) = e^{i\omega_dp} \bar{k}_d.
\end{align}
Since we assume the same operation is applied to the queries and keys, from now on, we will use $\mathbf{z}$ to refer to operations which act on both. In matrix form, this is given by $e^{i\omega_d p}$ can be represented as
\begin{equation}
    e^{i\omega_d p} \equiv
    \begin{bmatrix}
        \cos(\omega_dp) & -\sin (\omega_dp)\\ 
        \sin(\omega_dp) & \cos(\omega_dp)
    \end{bmatrix} = \mathbf{R}_{\omega_dp},
\end{equation}
where $\mathbf{R}_{\omega_dp_t}$ is a rotation matrix. While the rotation matrix is more intuitive, the complex exponential form will be useful for the mathematics in Section~\ref{sec:generality}, so we will alternate between the two. Recall, we use the convention that real valued queries and keys will have dimension $N$ and the complex interpretation will have dimension $D$.

One can represent the effect of RoPE as the application of a block diagonal of rotation matrices,
\begin{equation} \label{eq:RoPE}
RoPE(\mathbf{z},p) = \mathbf{R}_p \mathbf{z} = 
\begin{bmatrix}
\mathbf{R}_{p\omega_1} & \mathbf{0} & \cdots \\
\mathbf{0} & \mathbf{R}_{p\omega_2} & \cdots \\
\vdots & \vdots & \ddots
\end{bmatrix} 
\begin{pmatrix}
\mathbf{z}_1\\
\vdots
\\
\mathbf{z}_{D}
\end{pmatrix}
\equiv
\begin{bmatrix}
e^{ip\omega_1} & 0 & \cdots \\
0 & e^{ip\omega_2} & \cdots \\
\vdots & \vdots & \ddots
\end{bmatrix} 
\bar{\mathbf{z}}
,
\end{equation}
where $\mathbf{z}_d$ is a query pair. 
We introduce this block-diagonal form as it was the notation used in \citet{su2024roformer}. However, we will primarily stick to the index notation in Eq. \ref{eq:rope_index}.

\subsection{2D RoPE Embeddings}

RoPE is constrained to operate on (1D) sequences.
Motivated by the success of RoPE in language modeling, there have been growing efforts to extend it to multi-dimensional positions \cite{chen2024rotary,chu2024visionllama}, which we outline below.
We will use $M$ to refer to the dimensionality of the position, but will primarily focus on images, where $M=2$ and $p_i,p_j\in \mathbb{R}^2$.

\paragraph{Trivial 2D RoPE.}
One could trivially encode $\mathbf{p}=(p_x,p_y)$ using rotation matrices $\mathbf{R}_{\omega_dp_x}$, $\mathbf{R}_{\omega_dp_y}$:
    \begin{equation}\label{naive_rope}
        \varphi(\mathbf{z}_d, \mathbf{p}) = \mathbf{R}_{\omega_dp_x} \mathbf{R}_{\omega_dp_y}\mathbf{q}_d = \mathbf{R}_{\omega_d(p_x+p_y)} \mathbf{z}_d.
    \end{equation} 
However, in this case all positions with $p_x+p_y=c$ would get the same positional encoding $\mathbf{R}_{\omega_d c}$.

\paragraph{Axial RoPE.}
More practically, RoPE is extended to multiple dimensions by letting $x$ and $y$ act on different dimensions,
\begin{equation}
    \varphi(\mathbf{z}_d,\mathbf{p}) = \begin{bmatrix}
        \mathbf{R}_{p_x\omega_d} & \mathbf{0} \\
        \mathbf{0} & \mathbf{R}_{p_y\omega_d}
    \end{bmatrix} \mathbf{z}_d,
\end{equation}
where queries and keys are now split into four-dimensional vectors, $\mathbf{z}_d^\top = \begin{bmatrix}
    z^{(x)}_1,z^{(x)}_2,z^{(y)}_1,z^{(y)}_2
\end{bmatrix}$. 
The block-diagonal matrix can once again be viewed as a tensor of shape $N/2M\times M \times 2$, where $M$ is once again the dimensionality of position -- in this case $M=2$ for horizontal and vertical position. This gives the index notation
\begin{equation}\label{eq:modality}
    \varphi(\mathbf{z}_{d,m},p_m) = \mathbf{R}_{\omega_{d,m} p_m} \mathbf{z}_{d,m},
\end{equation}
for $m\in\{x,y\}$.
From a programming perspective, one can interpret this as a form of batched matrix multiplication.

\begin{figure}
    \centering
    \includegraphics[width=0.9\linewidth]{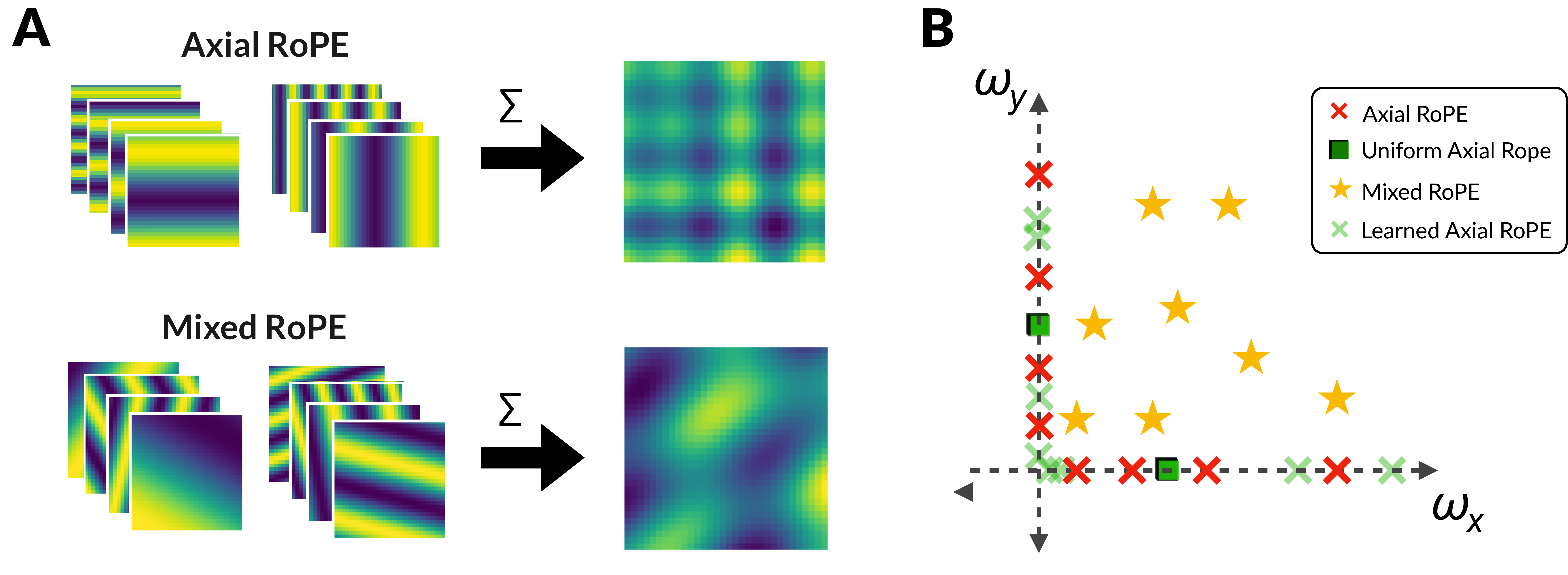}
    \caption{The attention patterns of Axial and Mixed RoPE. \textbf{A.} Each dimension pair in the query and key vectors is rotated based on the position creating an \textit{attention pattern}. The pixel value of the attention pattern is $\alpha(\mathbf{x}_q,\mathbf{x}_k, \mathbf{p}, \mathbf{0})$, where $\mathbf{p} = (i,j)$ -- the pixel location. On the left, the attention pattern of individual component-pairs in the embedding vector is shown and, on the right, the components are combined into the overall attention pattern for a randomly sampled query and key vector. \textbf{B.} Location of the rotations frequencies in 2D frequency space. Axial RoPE can only represent frequencies that lie on an axis resulting in the grid-like attention patterns. Unlike Axial RoPE, Mixed RoPE can assign different directions to each component-pair (\textbf{A} Bottom). When Axial RoPE uses fixed frequencies, the frequencies are spread exponentially. However, they can be implemented as learnable parameters. For Uniform RoPE, all frequencies are fixed to a single value for each axis.}
    \label{fig:att_pattern}
\end{figure}

While this method eliminates the symmetry, it treats $x$ and $y$ as independent. The result is a separable attention score of the form 
\begin{equation}
    \alpha_{i,j} = \alpha^{(x)}_{ij} + \alpha^{(y)}_{ij},
\end{equation}
where $\alpha^{(x)}_{ij}$ and $\alpha^{(y)}_{ij}$ are components of the attention score which depend only on $p_x$ and $p_y,$ respectively.
The frequencies are restricted to the axes, hence it is called Axial RoPE. This over-emphasizes horizontal and vertical relationships at the expense of oblique directions creating gridded patterns shown in Figure \ref{fig:att_pattern}.
To represent oblique patterns, the rotations would have to be performed along directions that contain both an $x$ and a $y$ component, i.\,e. frequencies that are not aligned on the axis in Figure\ref{fig:att_pattern}. These frequencies have been referred to as ``mixed frequencies''~\cite{heo2024rotary}.

\paragraph{Mixed RoPE: Learned mixed frequencies.}
The inclusion of mixed frequencies has empirically been shown to positively impact learning and generalization \cite{heo2024rotary}.
The naive approach in Eq. \ref{naive_rope} is only a problem when $x$ and $y$ rotate by the same frequency in every dimension. 
One could instead parameterize the frequencies with two separate frequencies in each dimension,
\begin{equation} \label{eq:LM_mixed}
    \varphi(\mathbf{z}_d) = \mathbf{R}_{\omega_{dx}x} \mathbf{R}_{\omega_{dy}y}\mathbf{q}_d = \mathbf{R}_{\omega_{dx} x+\omega_{dy}y} \mathbf{q}_d.
\end{equation}
By making the $\omega_x$ and $\omega_y$ parameters learnable, the attention pattern can learn mixed-frequency patterns by constructing a superposition of different diagonal patterns, as shown in Figure~\ref{fig:att_pattern}.

\paragraph{LieRE.}

Recently, RoPE has been interpreted through the lens of Lie algebras \cite{liu2025rethinking, ostmeier2024liere, schenck2025learning}. For an intuitive introduction to how Lie algebras appear, see Appendix \ref{app:linearflow}. Lie Rotary Position Encodings (LieRE)~\cite{ostmeier2024liere} extend Mixed RoPE by applying $N$-dimensional rotation matrices, rather than $2\times2$ matrices applied to pairs, using a linear combination of learned skew-symmetric Lie algebras,
\begin{equation}
    \varphi(\mathbf{z},p) = \exp(\mathcal{A}_x p_x + \mathcal{A}_y p_y)\mathbf{z},
\end{equation}
where $\exp$ is the matrix exponential, and the $\mathcal{A}$ terms are $N\times N$ skew-symmetric matrices -- which are Lie group \textit{generators} of a subgroup of $SO(N)$. Mathematically, LieRE is the most general rotary-based embedding method as skew-symmetric matrices are the generators of any $N$-D rotation. However, unlike the other two methods, LieRE is not guaranteed to be equivariant.

\section{The Generality of Learned RoPE and Mixed RoPE}\label{sec:generality}

While LieRE is motivated as generalizing RoPE to $M$-D rotations, in this section we will show that LieRE in one dimension can be learned by implementing RoPE with parameterized frequencies. For $M$-D positions, LieRE is not equivariant unless the generators commute. If the generators are required to commute, we show that LieRE can be re-expressed as Mixed RoPE. Thus, we conclude Mixed RoPE to be a general solution for $M$-D equivariant rotary embeddings. In this section, we will give informal proofs focused on high-level insights. 

\subsection{1D-LieRE is equivalent to 1D RoPE with learned frequencies}

In this section, we prove that any one-dimensional LieRE can be expressed as RoPE with parameterized rotation frequencies. 
Thus, we conclude RoPE to be a computationally efficient way of expressing a $D$-dimensional rotation,  i.\,e., 1D-LieRE. 
\vspace{4pt}
\begin{theorem}\label{th:generality}
Any 1D-LieRE can be parameterized by RoPE with learned frequencies.
\end{theorem}

To see why Proposition~\ref{th:generality} holds, suppose we have a 1D-LieRE embedding with a learned generator $\mathcal{A}$. By formulation, $\mathcal{A}$ is skew-symmetric, $\mathcal{A}^\top=-\mathcal{A}$. 
The positionally encoded attention between query $\mathbf{q}=\mathbf{W_q x}$ and $\mathbf{k}=\mathbf{W_k x}$ is
\begin{equation}
\alpha(x_i, x_j, p_i,p_j) = ( \exp(\mathcal{A}p_q)\mathbf{q})^\top \exp(\mathcal{A}p_k)\mathbf{k}.
\end{equation}
Any skew-symmetric matrix has an eigenvalue decomposition $\mathcal{A} = \mathbf{U} \mathbf{\Lambda}_\mathcal{I} \mathbf{U}^\top$ where $\mathbf{\Lambda}_\mathcal{I}$ is a diagonal matrix of purely imaginary (or zero) eigenvalues and $\mathbf{U}$ is a unitary matrix, $\mathbf{U}^\top\mathbf{U}=\mathbb{I}$. Moreover, the matrix-exponential of an eigenvalue decomposition simplifies to $\exp(\mathbf{U} \mathbf{\Lambda}_\mathcal{I} \mathbf{U}^\top) = \mathbf{U} \exp(\mathbf{\Lambda}_\mathcal{I}) \mathbf{U}^\top$. 
This allows us to express attention as
\begin{align}
\alpha(x_i, x_j, p_i,p_j) &= \mathbf{q} ^\top\mathbf{U} \exp(-p_q\mathbf{\Lambda}_\mathcal{I}) \mathbf{U}^\top \mathbf{U} \exp(p_k\mathbf{\Lambda}_\mathcal{I}) \mathbf{U}^\top\mathbf{k} \\&= \mathbf{q}'^\top \exp(-p_q\mathbf{\Lambda}_\mathcal{I}) \exp(p_k\mathbf{\Lambda}_\mathcal{I})\mathbf{k}',
\end{align}
where $\mathbf{q}' = \mathbf{W}'_q \mathbf{x}$ with $\mathbf{W}'_q = \mathbf{U}\mathbf{W}_q$, 
and 
$\mathbf{k}' = \mathbf{W}'_k \mathbf{\mathbf{x}}$ 
with $\mathbf{W}'_k = \mathbf{U}^\top \mathbf{W}_k$. Because the eigenvalue matrix is diagonal, the exponential is given by
\begin{equation}
    \exp(p\mathbf{\Lambda_{\mathcal{I}})} = \begin{bmatrix}
        e^{i\lambda_0p} & 0 & ... &0 \\
        0&e^{i\lambda_1p} & \ddots & \vdots\\
        \vdots & \ddots &  \ddots &0 \\
        0 & ... & 0 & e^{i\lambda_{N-1}p}
    \end{bmatrix}.
\end{equation}
Notice that this is the same as the complex formulation of RoPE defined in Eq. \ref{eq:RoPE}, where the eigenvalues of the generator correspond to the rotation frequencies of the rotation matrices. 
Thus, \textit{any} 1D-LieRE can be expressed as RoPE with learnable frequencies by absorbing the matrix of eigenvectors of $\mathcal{A}$ into the weight matrices $\mathbf{W}_\mathbf{q}$ and $\mathbf{W}_\mathbf{k}$. 
Since 1D-LieRE learns a rotation in $SO(D)$, RoPE can be seen as an efficient way to represent a rotation in $\mathbb{R}^{D}$. 

\subsection{Extending RoPE to more than one dimension}

While this proof works for 1D positions, it does not generalize to $M$-D without introducing extra inductive biases or giving up equivariance.
By imposing constraints on $\mathcal{A}_x$ and $\mathcal{A}_y$, we can categorize the other RoPE methods based on the assumptions made. 

\paragraph{Generators rotate independent subspaces.}
For example, one can impose the assumption that $p_x$ and $p_y$ rotate independent subspaces in $\mathbb{R}^N$. 
Mathematically, this assumption would imply that 
\begin{equation}
    \forall d\in [1,D]:  \lambda^{(x)}_d =0  \text{ or } \lambda^{(y)}_d =0,
\end{equation} 
where $\lambda^{(x)}_d$ and $\lambda^{(y)}_d$ are the eigenvalues of $\mathcal{A}_x$ and $\mathcal{A}_y$, respectively.
This is equivalent to rotating independent components of the query/key as done by Axial-RoPE. 

\paragraph{Commutative generators.}
For LieRE to be equivariant, we only need to ensure that the generators commute. If we make this assumption, then we arrive at Mixed RoPE.
\vspace{6pt}

\begin{theorem}
\label{th:liremixed}
Any $M$-dimensional LieRE with commutative generators can be parameterized by Mixed RoPE.
\end{theorem}

To see why Proposition~\ref{th:liremixed} holds, suppose we can diagonalize $\mathcal{A}_{x} = \mathbf{U}_{x}\mathbf{\Lambda}_{x}\mathbf{U}^\top_{x}$ and $\mathcal{A}_{y} = \mathbf{U}_{y}\mathbf{\Lambda}_{y}\mathbf{U}^\top_{y}$. If we take the assumption that $\mathcal{A}_{x}$ and $\mathcal{A}_{y}$ commute,
\begin{align}
    \mathcal{A}_{x} \mathcal{A}_{y} = \mathcal{A}_{y} \mathcal{A}_{x} ~~~ & \implies~~~[\mathcal{A}_{x}, \mathcal{A}_{y}] = [\mathcal{A}_{y}, \mathcal{A}_{x}] = 0.
\end{align}
$[\mathcal{A}_{x}, \mathcal{A}_{y}] = \mathcal{A}_{x} \mathcal{A}_{y} - \mathcal{A}_{y} \mathcal{A}_{x}$ is the Lie bracket. This implies that $\mathcal{A}_{x}$ and $\mathcal{A}_{y}$ are \textit{simultaneously diagonalizable} (Lemma~\ref{lemma:SimDiag}). Thus, commutativity implies that $\mathbf{U}_{x} = \mathbf{U}_{y} := \mathbf{U}$. We can write 
\begin{equation}
\mathbf{A} = \exp(\mathcal{A}_{x} p_x + \mathcal{A}_{y}p_y) = \mathbf{U} \exp(\mathbf{\Lambda}_{x}p_x + \mathbf{\Lambda}_{y}p_y) \mathbf{U}^\top,
\end{equation}
which leads to Eq. \ref{eq:LM_mixed}. 
Thus, Mixed RoPE forms a general solution for assumptions of commutativity, which is necessary for LieRE to be relative. 
This also mathematically shows Mixed RoPE to be the strict generalization of Axial RoPE.

In summary, learned frequency RoPE embeddings represent an efficient way of learning a much more general set of $SO(D)$ than is commonly believed. 
However, in order to generalize to higher dimensions while retaining their status as relative positional encodings, assumptions must be made. 
Mixed RoPE generalizes this to $M$-D positions, spanning the entire solution class for relative LieRE. 
Thus, LieRE-like methods with commutative $\mathcal{A}_x$ and $\mathcal{A}_y$ -- such as in \citet{schenck2025learning} -- are not more expressive than Mixed RoPE and any empirical differences in performance must be attributed to the learning dynamics due to different parameterizations. 
However, it remains unclear whether equivariance is the real reason for RoPE's success.

\begin{figure}
    \centering
    \includegraphics[width=1\linewidth]{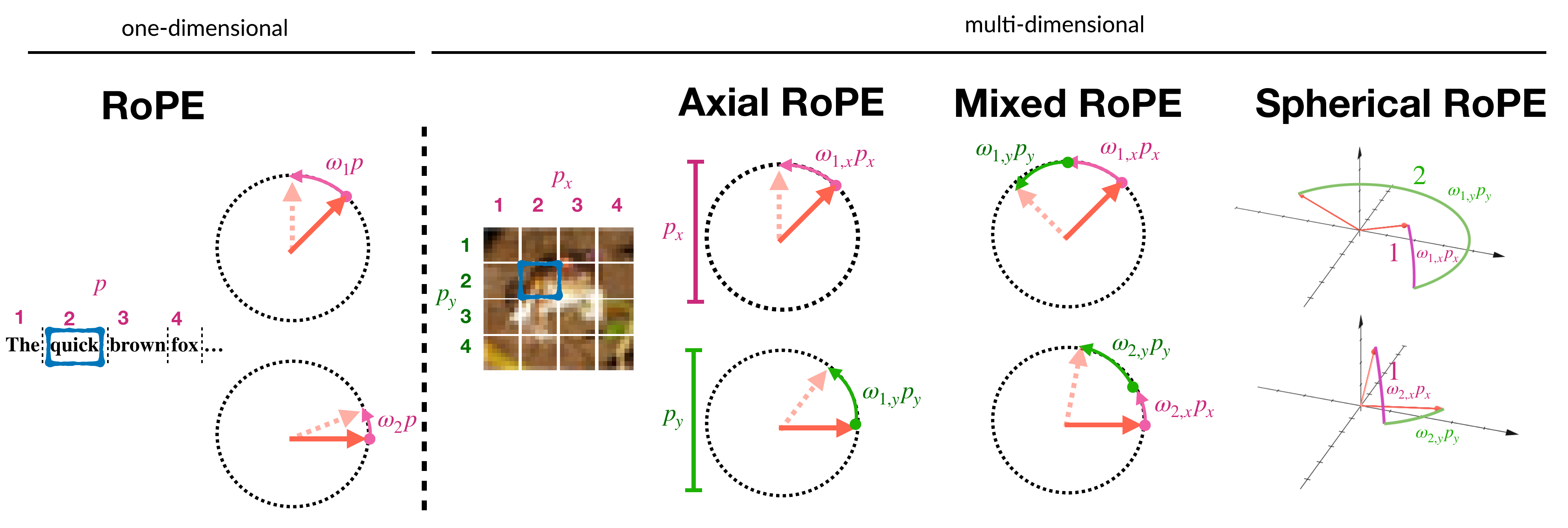}
    \caption{Diagram of each rotary embedding's effect on the subvector, $\mathbf{z}_d$. While Mixed RoPE affects 2D vector pairs, Spherical RoPE affects 3D vector triplets. Axial RoPE rotates independent dimensions for $p_x$, thus containing \textit{pairs of pairs}, or effectively quadruples. Each $\mathbf{z}$ contains $D$ sub-vectors rotating at different frequencies. While the order in which the rotations are applied does not matter for Axial or Mixed RoPE, order matters for Spherical RoPE. Explicitly, the triplet is first rotated around the axis associated with $p_x$ and then rotated around the axis associated with $p_y$.}
    \label{fig:enter-label}
\end{figure}

\section{Experiments}\label{sec:experiments}

When extending RoPE to more than one dimension, we must either constrain ourselves to commuting Lie algebras or give up relativity. We therefore ask the question: Why does RoPE work? Which properties should be preserved for generalizing RoPE to vision? 
To explore this question, we propose two new RoPE variants: \textit{Spherical RoPE}, which takes a non-commutative assumption, and \textit{Uniform-Frequency RoPE}, which uses a single fixed rotation frequency across all dimensions. 
Below we provide a high level outline the different embeddings. We compare the existing positional embedding methods APE \cite{dosovitskiy2020image}, Axial RoPE \cite{chu2024visionllama}, Mixed RoPE \cite{chen2024rotary}, and LieRE \cite{ostmeier2024liere} to these two new variants to understand whether equivariance, oblique directions or a variety of spatial frequencies are important features of PEs for vision. 

\paragraph{Spherical RoPE.}

We propose Spherical RoPE as a method between Mixed RoPE and LieRE that minimally changes 2D RoPE to break equivariance. 
Spherical RoPE embeds position as
\begin{equation} \label{spherical}
    \varphi(\mathbf{z}_d,\mathbf{p}) = \mathcal{Y}_{\omega_{dx}x} \mathcal{R}_{\omega_{dy}y}\mathbf{q}_d,
\end{equation}
where $\mathbf{q}_d\in\mathbb{R}^3$ is now a triplet instead of a pair, and $\mathcal{Y}$ is a block diagonal of $3\times 3$ \textit{yaw} matrices and $\mathcal{R}$ is a block diagonal of \textit{roll} matrices.
\begin{align}
    \mathcal{Y}_{\omega_{dx}x}  &= 
                                \begin{bmatrix}
                                \cos(\omega_{dx}x) & -\sin(\omega_{dx}x) & 0\\
                                \sin(\omega_{dx}x) & \cos(\omega_{dx}x) &0\\
                                0 & 0 & 1 \\
                                \end{bmatrix}
                                & & &
    \mathcal{R}_{\omega_{dy}y} &= \begin{bmatrix}
                                1 & 0 & 0 \\
                                0 & \cos(\omega_{dy}y) & -\sin(\omega_{dy}y) \\
                                0 & \sin(\omega_{dy}y) & \cos(\omega_{dy}y)
                                \end{bmatrix}.
\end{align}

Intuitively, rather than RoPE rotating around a circle, Spherical RoPE rotates around a sphere using Euler angles.

\begin{wraptable}{r}{0.625\textwidth}
\vspace{-0.5cm}
\caption{Table listing the properties of each of the rotary-based methods.}\label{tab:properties}
\centering
\footnotesize
\setlength{\tabcolsep}{2pt} %
\begin{tabular}{lccccccc}
\toprule
\textbf{Positional Encoding} & \textbf{Vision} & \textbf{Strictly} & \textbf{Oblique }  & \textbf{Requires} \\
& &\textbf{Equivariant} & \textbf{Directions} & \textbf{Learning} \\
\midrule
Rotary (RoPE)  \cite{su2024roformer}       & \xmark & \cmark & N/A & \xmark   \\
\midrule
Axial RoPE \cite{NDRoPE}         & \cmark & \cmark & \xmark & \xmark  \\
Mixed RoPE  \cite{heo2024rotary}       & \cmark & \cmark & \cmark & \cmark   \\
LieRE \cite{ostmeier2024liere}       & \cmark & \xmark & \cmark & \cmark   \\
\midrule
Spherical RoPE         & \cmark & \xmark & \cmark & \xmark  \\
Uniform RoPE         & \cmark & \cmark & \xmark & \xmark \\
\bottomrule
\end{tabular}
\end{wraptable}

Importantly, spherical rotations like LieRE are \textit{non-commutative} making them \textit{not equivariant}.
In fact, their generators are strictly \textit{non-commutative}, $\mathcal{A}_x\mathcal{A}_y \ne\mathcal{A}_y\mathcal{A}_x$. While non-commutativity does not mean Spherical RoPE is incapable of learning or approximating equivariance throughout the network, it is the component of LieRE removed by Mixed RoPE and works which enforce commutativity such as \citet{yu2025comropescalablerobustrotary} and \citet{schenck2025learning}.

We hypothesized Spherical RoPE to have a number of advantages. 
While Axial RoPE is unable to express oblique directions, Spherical RoPE can.
Like Axial RoPE, Spherical RoPE can use fixed frequencies making it computationally cheaper than LieRE and Mixed RoPE since sines and cosines of the frequencies can be precomputed. However, our main interest is that Spherical RoPE is comparable in terms of expressivity to Mixed and Axial RoPE while being non-equivariant.


\paragraph{Uniform-Frequency RoPE.}
For an initial evaluation on the impact of relative position, we propose Uniform-Frequency RoPE. 
For this method, we perform Axial RoPE with a single frequency shared across all rotation matrices. While still being relative, this serves as a more restricted version of RoPE. If this method performs significantly worse than other methods, it indicates the importance of having a range of frequencies. We implement uniform frequencies for Axial RoPE to gauge against relative importance of equivariance.

In one extreme, the rotation frequency could be zero resulting in no changes to the queries and keys. 
In the other extreme, the frequency could be set very high  resulting in large changes to the queries and keys. As a note, it is the frequency relative to the resolution of the image that is important. 
Frequencies higher than the sampling rate are equivalent to low frequencies. 
To ensure every position has a unique encoding, we fix the frequency to perform one rotation cycle across the entire image.

\paragraph{Datasets and architecture.}

We test the different PEs on CIFAR100 \cite{krizhevsky2009learning} and ImageNet \cite{russakovsky2015imagenet} using a standard Vision Transformer -- the ViT-S implementation from the timm~\cite{rw2019timm} library. For Learned APE, we use the baseline ViT-S which uses learned positional encodings rather than sinusoidal. We follow much of the DeiT-III training procedure proposed in \citet{touvron2022deit}. However, for ImageNet, we do not use dropout, MixUp, or CutMix as we observed that they significantly increase the number of epochs necessary for convergence. For ImageNet, we evaluate models trained after 200 epochs and 400 epochs for CIFAR100. We evaluate without any hyperparameter tuning directly on the validation sets. For further details on hyperparameters and experimental setup, see Appendix~\ref{app:hyperparameters}. Error bars were created using three models with different random seeds.

\paragraph{Generalization to larger image sizes.}

We also perform an experiment to test how well different PEs generalize across image sizes. Our approach to this experiment follows prior research~\cite{heo2024rotary, ostmeier2024liere}. The learned embeddings in Learned APE cannot be extrapolated, so we interpolate new embeddings when changing the number of patches. For RoPE embeddings, we take square dimensions and parameterize position such that the top-left corner of the image corresponds to $p_x=p_y=-\pi$ and the bottom-right corner correspond to $p_x=p_y=\pi$ with all other positions are evenly spread between the two for training. When increasing the image size, we extrapolate by scaling the range by the ratio of the new image size to the training image size while keeping the patch size constant.

\paragraph{Additional Evaluations}
Additional evaluations can be found in Appendix \ref{app:AddResults} including method speeds, experiments with smaller data splits, a segmentation task, and evaluation of the learned weights.

\section{Results} \label{sec:results}

To evaluate the importance of different properties of positional embeddings in vision transformers, we trained the same ViT with different positional embeddings on CIFAR100 and ImageNet-1K. We start by evaluating the models on images of the same resolution as during training. If equivariance is important, we would see Axial and Mixed RoPE to perform better than Spherical RoPE, which lacks strict equivariance. On the other hand, if oblique frequencies are important, then we would observe Mixed and Spherical RoPE to do better than Axial RoPE, which does not capture oblique directions.  We find that the lack of equivariance does not hinder Spherical RoPE. It outperforms Axial RoPE and performs comparably to Mixed RoPE. Moreover, we would expect equivariant methods to be especially effective in the low data regime. However, in Appendix \ref{app:AddResults} we observe Learned Spherical RoPE performs the best despite its lack of inductive bias. This suggests that the benefits to performance and generalization on ImageNet for Mixed RoPE may be due to its extra parameters. However, Axial RoPE and Uniform RoPE perform significantly worse suggesting oblique directions to be more important than equivariance. 

\begin{figure}[t]
\centering
\begin{minipage}[t]{0.50\textwidth}
\vspace{-4.7cm} 
\captionof{table}{
    Performance comparison (top-1 accuracy) across datasets and methods.
}\label{tab:Results}
\centering
\resizebox{\linewidth}{!}{
\begin{tabular}{lrrr}
    \toprule
     & \multicolumn{2}{c}{\textbf{Top-1 Accuracy (\%)}}  \\
    \cmidrule{2-3}
    \textbf{Fixed Encoding}
    & \textbf{CIFAR100} & \textbf{ImageNet}  \\
    \midrule
    Learned APE & 64.2{\tiny$\pm0.9$} & 72.7 {\tiny $\pm 0.1$}  \\
    Axial RoPE & 72.1{\tiny$\pm0.6$} & 75.6 {\tiny $\pm 0.2$}  \\
    Uniform RoPE (Our Ablation) & 70.5{\tiny$\pm 0.2$} & 74.9 {\tiny $\pm 0.3$} \\
    Spherical RoPE (Our Ablation) & \textbf{73.2}{\tiny$\pm0.4$} & \textbf{76.4} {\tiny $\pm 0.3$}\\
    \midrule
    \midrule
    \textbf{Learned Encoding}\\
    \midrule
    Learned Axial RoPE & 72.9{\tiny$\pm0.6$} & 75.7 {\tiny $\pm 0.4$}\\
    LieRE & 73.1{\tiny$\pm0.2$}  \\
    Mixed RoPE & \textbf{74.7}{\tiny$\pm0.3$} & \textbf{77.4 {\tiny$\pm 0.1$}}  \\
    Learned Spherical RoPE (Our Ablation) & 74.1{\tiny$\pm0.4$} &  \textbf{77.4}  {\tiny$\pm 0.2$}  \\
    \bottomrule
\end{tabular}
}
\end{minipage}%
\hfill
\begin{minipage}[t]{0.45\textwidth}
\centering
\includegraphics[width=\textwidth]{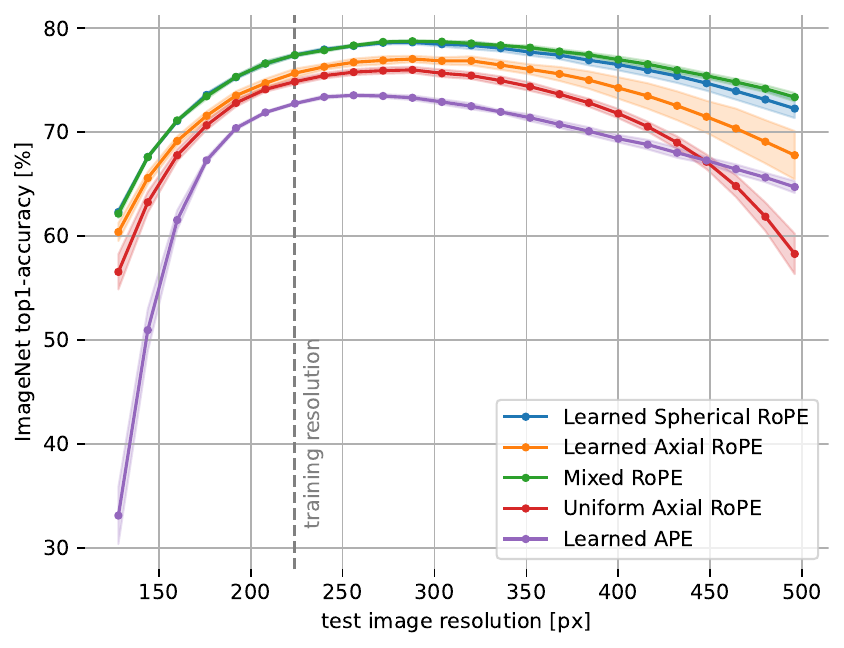}
\caption{Dependence of accuracy on image resolution for ViT-S with various positional embedding methods on ImageNet-1k. Error bars reflect the standard deviation across three models trained with different seeds.}
\label{fig:resolution}
\end{minipage}
\end{figure}

When comparing with absolute positional encodings, we observe that all forms of RoPE perform better than learned APE (Table~\ref{tab:Results}). This includes Uniform RoPE, the variant that uses only a single frequency. Moreover, all forms of RoPE using diverse frequencies outperform Uniform RoPE and have similar performance (whether they are learned or not), suggesting that diversity of frequencies is important. Spherical RoPE adheres much closer to the vectorized implementation of other RoPE methods than LieRE. As our goal was primarily to identify the most impactful properties of $M$-D RoPE and not maximize accuracy, none of our conclusions depend on precise performance numbers for LieRE.

Last, we asked how well different PEs generalize across image sizes. Equivariance is often thought to aid model  generalization. However, when evaluating each model using higher resolutions images, i.\,e. increasing the number of patches, we found Spherical RoPE to be the most effective method (Figure~\ref{fig:resolution}), suggesting equivariance may not be the reason for RoPE's generalization.

\section{Discussion}
Because we see very little variation between Spherical RoPE and Mixed RoPE, we conclude that equivariance is only a minor contributor to the increased performance seen by RoPE for vision. In fact, Spherical RoPE appeared to extrapolate to higher resolutions better than Axial RoPE. This could suggest that oblique frequencies are important for extrapolation. However, extrapolation is only done on short length scales, so this may not hold in language.

There are two important differences between vision and language transformers: context length and patch variation. Where LLMs have on the order of 128K context windows \cite{grattafiori2024llama}, vision transformers only have $16\times 16$ patches. Moreover, patches have more variation as tokens than language tokens, thus allowing the content embeddings to store information about the relevance of oblique directions. Because the context size is small, we hypothesize that there could be methods that perform better than Mixed RoPE and are more general than LieRE for vision. While LieRE was proposed with skew-symmetric generators to generalize RoPE to $N$-D rotations, Lie algebras do not have to be skew-symmetric. The skew-symmetry is important for maintaining numerical stability over long contexts~\cite{su2024roformer}. However, skew-symmetry also results in Proposition 3.1 of \citet{barbero2024round}, which proves RoPE to be non-local. Since the context size is small for images, numerical stability is likely not an issue, thus freeing the space of Lie algebras available to us -- including Lie algebras that encourage locality.

We observe a decrease in generalization when using uniform frequencies. This finding qualifies \citet{barbero2024round}'s hypothesis that the various semantic lengths contribute to RoPE's performance. However, Uniform RoPE outperformed learned APE, suggesting the reason why RoPE performs well is not among the properties we tested. We speculate this could be a flaw in additive positional embeddings themselves; additive methods create a trade-off between the magnitude of position and content -- forcing tokens that vary significantly with position to have lower magnitude to be closer to the origin. 

\section{Conclusion}
We conclude that Mixed RoPE is a very general solution for $M$-D data if equivariance is a necessity. However, we see little evidence that strict relative positional bias is impactful for vision transformers. However, RoPE methods have still been found to greatly improve performance in ViTs. Thus, we conclude that evidence suggests that RoPE does not \textit{need} strict equivariance constraints to boost performance over APE methods.

\section{Acknowledgments}

This project has received funding from the European Research Council (ERC) under the European Union’s Horizon Europe research and innovation programme (Grant agreement No. 101041669). We gratefully acknowledge the computing time granted by the Resource Allocation Board and provided on the supercomputer Emmy/Grete at NHR-Nord@Göttingen as part of the NHR infrastructure (project nim00012). 

\bibliographystyle{plainnat} 
\bibliography{biblio}  

@article{vaswani2017attention,
  title={Attention is all you need},
  author={Vaswani, Ashish and Shazeer, Noam and Parmar, Niki and Uszkoreit, Jakob and Jones, Llion and Gomez, Aidan N and Kaiser, {\L}ukasz and Polosukhin, Illia},
  journal={Advances in neural information processing systems},
  volume={30},
  year={2017}
}

@article{bahdanau2014neural,
  title={Neural machine translation by jointly learning to align and translate},
  author={Bahdanau, Dzmitry and Cho, Kyunghyun and Bengio, Yoshua},
  journal={arXiv preprint arXiv:1409.0473},
  year={2014}
}

@article{xu2018powerful,
  title={How powerful are graph neural networks?},
  author={Xu, Keyulu and Hu, Weihua and Leskovec, Jure and Jegelka, Stefanie},
  journal={arXiv preprint arXiv:1810.00826},
  year={2018}
}

@article{kreuzer2021rethinking,
  title={Rethinking graph transformers with spectral attention},
  author={Kreuzer, Devin and Beaini, Dominique and Hamilton, Will and L{\'e}tourneau, Vincent and Tossou, Prudencio},
  journal={Advances in Neural Information Processing Systems},
  volume={34},
  pages={21618--21629},
  year={2021}
}

@inproceedings{yun2019cutmix,
  title={Cutmix: Regularization strategy to train strong classifiers with localizable features},
  author={Yun, Sangdoo and Han, Dongyoon and Oh, Seong Joon and Chun, Sanghyuk and Choe, Junsuk and Yoo, Youngjoon},
  booktitle={Proceedings of the IEEE/CVF international conference on computer vision},
  pages={6023--6032},
  year={2019}
}

@article{choromanski2020rethinking,
  title={Rethinking attention with performers},
  author={Choromanski, Krzysztof and Likhosherstov, Valerii and Dohan, David and Song, Xingyou and Gane, Andreea and Sarlos, Tamas and Hawkins, Peter and Davis, Jared and Mohiuddin, Afroz and Kaiser, Lukasz and others},
  journal={arXiv preprint arXiv:2009.14794},
  year={2020}
}

@misc{meta_llama4_2025,
  author       = {Meta AI},
  title        = {LLaMA 4: Multimodal Language Models},
  year         = {2025},
  url          = {https://ai.meta.com/blog/llama-4-multimodal-intelligence/},
  note         = {Accessed: 2025-05-15}
}

@article{lee2021fnet,
  title={Fnet: Mixing tokens with fourier transforms},
  author={Lee-Thorp, James and Ainslie, Joshua and Eckstein, Ilya and Ontanon, Santiago},
  journal={arXiv preprint arXiv:2105.03824},
  year={2021}
}

@article{yang2025rope,
  title={Rope to Nope and Back Again: A New Hybrid Attention Strategy},
  author={Yang, Bowen and Venkitesh, Bharat and Talupuru, Dwarak and Lin, Hangyu and Cairuz, David and Blunsom, Phil and Locatelli, Acyr},
  journal={arXiv preprint arXiv:2501.18795},
  year={2025}
}

@inproceedings{elsayed2020revisiting,
  title={Revisiting spatial invariance with low-rank local connectivity},
  author={Elsayed, Gamaleldin and Ramachandran, Prajit and Shlens, Jonathon and Kornblith, Simon},
  booktitle={International Conference on Machine Learning},
  pages={2868--2879},
  year={2020},
  organization={PMLR}
}

@misc{yu2025comropescalablerobustrotary,
      title={ComRoPE: Scalable and Robust Rotary Position Embedding Parameterized by Trainable Commuting Angle Matrices}, 
      author={Hao Yu and Tangyu Jiang and Shuning Jia and Shannan Yan and Shunning Liu and Haolong Qian and Guanghao Li and Shuting Dong and Huaisong Zhang and Chun Yuan},
      year={2025},
      eprint={2506.03737},
      archivePrefix={arXiv},
      primaryClass={cs.CV},
      url={https://arxiv.org/abs/2506.03737}, 
}

@article{islam2025platonic,
  title={Platonic Transformers: A Solid Choice For Equivariance},
  author={Islam, Mohammad Mohaiminul and Anand, Rishabh and Wessels, David R and de Kruiff, Friso and Kuipers, Thijs P and Ying, Rex and S{\'a}nchez, Clara I and Vadgama, Sharvaree and B{\"o}kman, Georg and Bekkers, Erik J},
  journal={arXiv preprint arXiv:2510.03511},
  year={2025}
}

@article{Everingham15,
  author    = {Mark Everingham and
               S. M. Ali Eslami and
               Luc Van Gool and
               Christopher K. I. Williams and
               John Winn and
               Andrew Zisserman},
  title     = {The {Pascal} Visual Object Classes Challenge: {A} Retrospective},
  journal   = {International Journal of Computer Vision},
  volume    = {111},
  number    = {1},
  pages     = {98--136},
  year      = {2015},
  doi       = {10.1007/s11263-014-0733-5}
}

@article{chu2024visionllama,
  title={Visionllama: A unified llama interface for vision tasks},
  author={Chu, Xiangxiang and Su, Jianlin and Zhang, Bo and Shen, Chunhua},
  journal={arXiv e-prints},
  pages={arXiv--2403},
  year={2024}
}

@inproceedings{touvron2022deit,
  title={Deit iii: Revenge of the vit},
  author={Touvron, Hugo and Cord, Matthieu and J{\'e}gou, Herv{\'e}},
  booktitle={European conference on computer vision},
  pages={516--533},
  year={2022},
  organization={Springer}
}

@article{zhang2017mixup,
  title={mixup: Beyond empirical risk minimization},
  author={Zhang, Hongyi and Cisse, Moustapha and Dauphin, Yann N and Lopez-Paz, David},
  journal={arXiv preprint arXiv:1710.09412},
  year={2017}
}

@article{you2019large,
  title={Large batch optimization for deep learning: Training bert in 76 minutes},
  author={You, Yang and Li, Jing and Reddi, Sashank and Hseu, Jonathan and Kumar, Sanjiv and Bhojanapalli, Srinadh and Song, Xiaodan and Demmel, James and Keutzer, Kurt and Hsieh, Cho-Jui},
  journal={arXiv preprint arXiv:1904.00962},
  year={2019}
}

@article{wang2024length,
  title={Length generalization of causal transformers without position encoding},
  author={Wang, Jie and Ji, Tao and Wu, Yuanbin and Yan, Hang and Gui, Tao and Zhang, Qi and Huang, Xuanjing and Wang, Xiaoling},
  journal={arXiv preprint arXiv:2404.12224},
  year={2024}
}

@article{shamshad2023transformers,
  title={Transformers in medical imaging: A survey},
  author={Shamshad, Fahad and Khan, Salman and Zamir, Syed Waqas and Khan, Muhammad Haris and Hayat, Munawar and Khan, Fahad Shahbaz and Fu, Huazhu},
  journal={Medical image analysis},
  volume={88},
  pages={102802},
  year={2023},
  publisher={Elsevier}
}

@article{park2022grpe,
  title={Grpe: Relative positional encoding for graph transformer},
  author={Park, Wonpyo and Chang, Woonggi and Lee, Donggeon and Kim, Juntae and Hwang, Seung-won},
  journal={arXiv preprint arXiv:2201.12787},
  year={2022}
}

@article{jumper2021highly,
  title={Highly accurate protein structure prediction with AlphaFold},
  author={Jumper, John and Evans, Richard and Pritzel, Alexander and Green, Tim and Figurnov, Michael and Ronneberger, Olaf and Tunyasuvunakool, Kathryn and Bates, Russ and {\v{Z}}{\'\i}dek, Augustin and Potapenko, Anna and others},
  journal={nature},
  volume={596},
  number={7873},
  pages={583--589},
  year={2021},
  publisher={Nature Publishing Group}
}

@inproceedings{you2019position,
  title={Position-aware graph neural networks},
  author={You, Jiaxuan and Ying, Rex and Leskovec, Jure},
  booktitle={International conference on machine learning},
  pages={7134--7143},
  year={2019},
  organization={PMLR}
}

@article{wei2025videorope,
  title={VideoRoPE: What Makes for Good Video Rotary Position Embedding?},
  author={Wei, Xilin and Liu, Xiaoran and Zang, Yuhang and Dong, Xiaoyi and Zhang, Pan and Cao, Yuhang and Tong, Jian and Duan, Haodong and Guo, Qipeng and Wang, Jiaqi and others},
  journal={arXiv preprint arXiv:2502.05173},
  year={2025}
}

@article{wang2024qwen2,
  title={Qwen2-vl: Enhancing vision-language model's perception of the world at any resolution},
  author={Wang, Peng and Bai, Shuai and Tan, Sinan and Wang, Shijie and Fan, Zhihao and Bai, Jinze and Chen, Keqin and Liu, Xuejing and Wang, Jialin and Ge, Wenbin and others},
  journal={arXiv preprint arXiv:2409.12191},
  year={2024}
}

@article{liu2025vrope,
  title={VRoPE: Rotary Position Embedding for Video Large Language Models},
  author={Liu, Zikang and Guo, Longteng and Tang, Yepeng and Cai, Junxian and Ma, Kai and Chen, Xi and Liu, Jing},
  journal={arXiv preprint arXiv:2502.11664},
  year={2025}
}

@article{schenck2025learning,
  title={Learning the RoPEs: Better 2D and 3D Position Encodings with STRING},
  author={Schenck, Connor and Reid, Isaac and Jacob, Mithun George and Bewley, Alex and Ainslie, Joshua and Rendleman, David and Jain, Deepali and Sharma, Mohit and Dubey, Avinava and Wahid, Ayzaan and others},
  journal={arXiv preprint arXiv:2502.02562},
  year={2025}
}

@article{chi2022dissecting,
  title={Dissecting transformer length extrapolation via the lens of receptive field analysis},
  author={Chi, Ta-Chung and Fan, Ting-Han and Rudnicky, Alexander I and Ramadge, Peter J},
  journal={arXiv preprint arXiv:2212.10356},
  year={2022}
}

@article{rahimi2007random,
  title={Random features for large-scale kernel machines},
  author={Rahimi, Ali and Recht, Benjamin},
  journal={Advances in neural information processing systems},
  volume={20},
  year={2007}
}

@article{brehmer2024does,
  title={Does equivariance matter at scale?},
  author={Brehmer, Johann and Behrends, S{\"o}nke and de Haan, Pim and Cohen, Taco},
  journal={arXiv preprint arXiv:2410.23179},
  year={2024}
}

@article{pham2020relative,
  title={Relative positional encoding for speech recognition and direct translation},
  author={Pham, Ngoc-Quan and Ha, Thanh-Le and Nguyen, Tuan-Nam and Nguyen, Thai-Son and Salesky, Elizabeth and St{\"u}ker, Sebastian and Niehues, Jan and Waibel, Alexander},
  journal={arXiv preprint arXiv:2005.09940},
  year={2020}
}

@article{team2024gemma,
  title={Gemma: Open models based on gemini research and technology},
  author={Team, Gemma and Mesnard, Thomas and Hardin, Cassidy and Dadashi, Robert and Bhupatiraju, Surya and Pathak, Shreya and Sifre, Laurent and Rivi{\`e}re, Morgane and Kale, Mihir Sanjay and Love, Juliette and others},
  journal={arXiv preprint arXiv:2403.08295},
  year={2024}
}

@article{liu2024deepseek,
  title={Deepseek-v3 technical report},
  author={Liu, Aixin and Feng, Bei and Xue, Bing and Wang, Bingxuan and Wu, Bochao and Lu, Chengda and Zhao, Chenggang and Deng, Chengqi and Zhang, Chenyu and Ruan, Chong and others},
  journal={arXiv preprint arXiv:2412.19437},
  year={2024}
}

@article{dao2022flashattention,
  title={Flashattention: Fast and memory-efficient exact attention with io-awareness},
  author={Dao, Tri and Fu, Dan and Ermon, Stefano and Rudra, Atri and R{\'e}, Christopher},
  journal={Advances in neural information processing systems},
  volume={35},
  pages={16344--16359},
  year={2022}
}

@article{angelotti2023hype,
  title={HyPE: Attention with Hyperbolic Biases for Relative Positional Encoding},
  author={Angelotti, Giorgio},
  journal={arXiv preprint arXiv:2310.19676},
  year={2023}
}

@article{abramson2024accurate,
  title={Accurate structure prediction of biomolecular interactions with AlphaFold 3},
  author={Abramson, Josh and Adler, Jonas and Dunger, Jack and Evans, Richard and Green, Tim and Pritzel, Alexander and Ronneberger, Olaf and Willmore, Lindsay and Ballard, Andrew J and Bambrick, Joshua and others},
  journal={Nature},
  volume={630},
  number={8016},
  pages={493--500},
  year={2024},
  publisher={Nature Publishing Group UK London}
}

@article{brandstetter2021geometric,
  title={Geometric and physical quantities improve e (3) equivariant message passing},
  author={Brandstetter, Johannes and Hesselink, Rob and van der Pol, Elise and Bekkers, Erik J and Welling, Max},
  journal={arXiv preprint arXiv:2110.02905},
  year={2021}
}

@inproceedings{schutt2021equivariant,
  title={Equivariant message passing for the prediction of tensorial properties and molecular spectra},
  author={Sch{\"u}tt, Kristof and Unke, Oliver and Gastegger, Michael},
  booktitle={International Conference on Machine Learning},
  pages={9377--9388},
  year={2021},
  organization={PMLR}
}

@article{chu2021conditional,
  title={Conditional positional encodings for vision transformers},
  author={Chu, Xiangxiang and Tian, Zhi and Zhang, Bo and Wang, Xinlong and Shen, Chunhua},
  journal={arXiv preprint arXiv:2102.10882},
  year={2021}
}

@inproceedings{gilmer2017neural,
  title={Neural message passing for quantum chemistry},
  author={Gilmer, Justin and Schoenholz, Samuel S and Riley, Patrick F and Vinyals, Oriol and Dahl, George E},
  booktitle={International conference on machine learning},
  pages={1263--1272},
  year={2017},
  organization={PMLR}
}

@article{kofinas2021roto,
  title={Roto-translated local coordinate frames for interacting dynamical systems},
  author={Kofinas, Miltiadis and Nagaraja, Naveen and Gavves, Efstratios},
  journal={Advances in Neural Information Processing Systems},
  volume={34},
  pages={6417--6429},
  year={2021}
}

@article{li2020distance,
  title={Distance encoding: Design provably more powerful neural networks for graph representation learning},
  author={Li, Pan and Wang, Yanbang and Wang, Hongwei and Leskovec, Jure},
  journal={Advances in Neural Information Processing Systems},
  volume={33},
  pages={4465--4478},
  year={2020}
}

@article{schmidhuber1992learning,
  title={Learning to control fast-weight memories: An alternative to dynamic recurrent networks},
  author={Schmidhuber, J{\"u}rgen},
  journal={Neural Computation},
  volume={4},
  number={1},
  pages={131--139},
  year={1992},
  publisher={MIT Press One Rogers Street, Cambridge, MA 02142-1209, USA journals-info~…}
}

@article{velivckovic2017graph,
  title={Graph attention networks},
  author={Veli{\v{c}}kovi{\'c}, Petar and Cucurull, Guillem and Casanova, Arantxa and Romero, Adriana and Lio, Pietro and Bengio, Yoshua},
  journal={arXiv preprint arXiv:1710.10903},
  year={2017}
}

@inproceedings{he2023sheaf,
  title={Sheaf-based positional encodings for graph neural networks},
  author={He, Yu and Bodnar, Cristian and Lio, Pietro},
  booktitle={NeurIPS 2023 Workshop on Symmetry and Geometry in Neural Representations},
  volume={9},
  year={2023}
}

@inproceedings{katharopoulos2020transformers,
  title={Transformers are rnns: Fast autoregressive transformers with linear attention},
  author={Katharopoulos, Angelos and Vyas, Apoorv and Pappas, Nikolaos and Fleuret, Fran{\c{c}}ois},
  booktitle={International conference on machine learning},
  pages={5156--5165},
  year={2020},
  organization={PMLR}
}

@article{su2024roformer,
  title={Roformer: Enhanced transformer with rotary position embedding},
  author={Su, Jianlin and Ahmed, Murtadha and Lu, Yu and Pan, Shengfeng and Bo, Wen and Liu, Yunfeng},
  journal={Neurocomputing},
  volume={568},
  pages={127063},
  year={2024},
  publisher={Elsevier}
}

@misc{rope-eleutherai,
  title = {Rotary Embeddings: A Relative Revolution},
  author = {Biderman, Stella and Black, Sid and Foster, Charles and Gao, Leo and Hallahan, Eric and He, Horace and Wang, Ben and Wang, Phil},
  howpublished = {\url{blog.eleuther.ai/}},
  note = {[Online; accessed ]},
  year = {2021}
}

@article{jiang2024mixtral,
  title={Mixtral of experts},
  author={Jiang, Albert Q and Sablayrolles, Alexandre and Roux, Antoine and Mensch, Arthur and Savary, Blanche and Bamford, Chris and Chaplot, Devendra Singh and Casas, Diego de las and Hanna, Emma Bou and Bressand, Florian and others},
  journal={arXiv preprint arXiv:2401.04088},
  year={2024}
}

@inproceedings{liu2021swin,
  title={Swin transformer: Hierarchical vision transformer using shifted windows},
  author={Liu, Ze and Lin, Yutong and Cao, Yue and Hu, Han and Wei, Yixuan and Zhang, Zheng and Lin, Stephen and Guo, Baining},
  booktitle={Proceedings of the IEEE/CVF international conference on computer vision},
  pages={10012--10022},
  year={2021}
}

@article{press2021train,
  title={Train short, test long: Attention with linear biases enables input length extrapolation},
  author={Press, Ofir and Smith, Noah A and Lewis, Mike},
  journal={arXiv preprint arXiv:2108.12409},
  year={2021}
}

@book{Griffiths,
    author = {Griffiths, D.J.},
    title = {{Introduction to Quantum Mechanics}},
    publisher = {CUP},
    year = {2018}
}

@article{shaw2018self,
  title={Self-attention with relative position representations},
  author={Shaw, Peter and Uszkoreit, Jakob and Vaswani, Ashish},
  journal={arXiv preprint arXiv:1803.02155},
  year={2018}
}

@article{raffel2020exploring,
  title={Exploring the limits of transfer learning with a unified text-to-text transformer},
  author={Raffel, Colin and Shazeer, Noam and Roberts, Adam and Lee, Katherine and Narang, Sharan and Matena, Michael and Zhou, Yanqi and Li, Wei and Liu, Peter J},
  journal={Journal of machine learning research},
  volume={21},
  number={140},
  pages={1--67},
  year={2020}
}

@article{DBLP:journals/corr/abs-2003-09229,
  author       = {Xuanqing Liu and
                  Hsiang{-}Fu Yu and
                  Inderjit S. Dhillon and
                  Cho{-}Jui Hsieh},
  title        = {Learning to Encode Position for Transformer with Continuous Dynamical
                  Model},
  journal      = {CoRR},
  volume       = {abs/2003.09229},
  year         = {2020},
  url          = {https://arxiv.org/abs/2003.09229},
  eprinttype    = {arXiv},
  eprint       = {2003.09229},
  bibsource    = {dblp computer science bibliography, https://dblp.org}
}

@article{grattafiori2024llama,
  title={The llama 3 herd of models},
  author={Grattafiori, Aaron and Dubey, Abhimanyu and Jauhri, Abhinav and Pandey, Abhinav and Kadian, Abhishek and Al-Dahle, Ahmad and Letman, Aiesha and Mathur, Akhil and Schelten, Alan and Vaughan, Alex and others},
  journal={arXiv preprint arXiv:2407.21783},
  year={2024}
}

@article{liu2023molrope,
  title={MolRoPE-BERT: An enhanced molecular representation with Rotary Position Embedding for molecular property prediction},
  author={Liu, Yunwu and Zhang, Ruisheng and Li, Tongfeng and Jiang, Jing and Ma, Jun and Wang, Ping},
  journal={Journal of Molecular Graphics and Modelling},
  volume={118},
  pages={108344},
  year={2023},
  publisher={Elsevier}
}

@inproceedings{heo2024rotary,
  title={Rotary position embedding for vision transformer},
  author={Heo, Byeongho and Park, Song and Han, Dongyoon and Yun, Sangdoo},
  booktitle={European Conference on Computer Vision},
  pages={289--305},
  year={2024},
  organization={Springer}
}

@article{dosovitskiy2020image,
  title={An image is worth 16x16 words: Transformers for image recognition at scale},
  author={Dosovitskiy, Alexey and Beyer, Lucas and Kolesnikov, Alexander and Weissenborn, Dirk and Zhai, Xiaohua and Unterthiner, Thomas and Dehghani, Mostafa and Minderer, Matthias and Heigold, Georg and Gelly, Sylvain and others},
  journal={arXiv preprint arXiv:2010.11929},
  year={2020}
}

@article{barbero2024round,
  title={Round and Round We Go! What makes Rotary Positional Encodings useful?},
  author={Barbero, Federico and Vitvitskyi, Alex and Perivolaropoulos, Christos and Pascanu, Razvan and Veli{\v{c}}kovi{\'c}, Petar},
  journal={arXiv preprint arXiv:2410.06205},
  year={2024}
}

@inproceedings{chen2024rotary,
  title={What rotary position embedding can tell us: Identifying query and key weights corresponding to basic syntactic or high-level semantic information},
  author={Chen, Yiting and Yan, Junchi},
  booktitle={The Thirty-eighth Annual Conference on Neural Information Processing Systems},
  year={2024}
}

@article{kazemnejad2023impact,
  title={The impact of positional encoding on length generalization in transformers},
  author={Kazemnejad, Amirhossein and Padhi, Inkit and Natesan Ramamurthy, Karthikeyan and Das, Payel and Reddy, Siva},
  journal={Advances in Neural Information Processing Systems},
  volume={36},
  pages={24892--24928},
  year={2023}
}

@article{ostmeier2024liere,
  title={Liere: Generalizing rotary position encodings},
  author={Ostmeier, Sophie and Axelrod, Brian and Moseley, Michael E and Chaudhari, Akshay and Langlotz, Curtis},
  journal={arXiv preprint arXiv:2406.10322},
  year={2024}
}

@article{haviv2022transformer,
  title={Transformer language models without positional encodings still learn positional information},
  author={Haviv, Adi and Ram, Ori and Press, Ofir and Izsak, Peter and Levy, Omer},
  journal={arXiv preprint arXiv:2203.16634},
  year={2022}
}

@article{bekkers2023fast,
  title={Fast, Expressive SE $(n) $ Equivariant Networks through Weight-Sharing in Position-Orientation Space},
  author={Bekkers, Erik J and Vadgama, Sharvaree and Hesselink, Rob D and Van der Linden, Putri A and Romero, David W},
  journal={arXiv preprint arXiv:2310.02970},
  year={2023}
}

@article{wessels2024grounding,
  title={Grounding continuous representations in geometry: Equivariant neural fields},
  author={Wessels, David R and Knigge, David M and Papa, Samuele and Valperga, Riccardo and Vadgama, Sharvaree and Gavves, Efstratios and Bekkers, Erik J},
  journal={arXiv preprint arXiv:2406.05753},
  year={2024}
}

@article{knigge2024space,
  title={Space-time continuous pde forecasting using equivariant neural fields},
  author={Knigge, David and Wessels, David and Valperga, Riccardo and Papa, Samuele and Sonke, Jan-Jakob and Bekkers, Erik and Gavves, Efstratios},
  journal={Advances in Neural Information Processing Systems},
  volume={37},
  pages={76553--76577},
  year={2024}
}

@article{russakovsky2015imagenet,
  title={Imagenet large scale visual recognition challenge},
  author={Russakovsky, Olga and Deng, Jia and Su, Hao and Krause, Jonathan and Satheesh, Sanjeev and Ma, Sean and Huang, Zhiheng and Karpathy, Andrej and Khosla, Aditya and Bernstein, Michael and others},
  journal={International journal of computer vision},
  volume={115},
  pages={211--252},
  year={2015},
  publisher={Springer}
}

@misc{krizhevsky2009learning,
  title={Learning multiple layers of features from tiny images},
  author={Krizhevsky, Alex and Hinton, Geoffrey and others},
  year={2009},
  publisher={Toronto, ON, Canada}
}

@article{kanerva2009hyperdimensional,
  title={Hyperdimensional computing: An introduction to computing in distributed representation with high-dimensional random vectors},
  author={Kanerva, Pentti},
  journal={Cognitive computation},
  volume={1},
  pages={139--159},
  year={2009},
  publisher={Springer}
}

@article{liu2025rethinking,
  title={Rethinking RoPE: A Mathematical Blueprint for N-dimensional Positional Encoding},
  author={Liu, Haiping and Zhou, Hongpeng},
  journal={arXiv preprint arXiv:2504.06308},
  year={2025}
}

@misc{NDRoPE,
  title={Transformer Upgrade Road: 4. Rotating position coding of two-dimensional position},
  author={Su, Jianlin},
  year={2021},
  month={May},
  url={https://kexue.fm/archives/8397},
  note={Accessed: 2025-04-28}
}

@article{kymn2024computing,
  title={Computing with residue numbers in high-dimensional representation},
  author={Kymn, Christopher J and Kleyko, Denis and Frady, E Paxon and Bybee, Connor and Kanerva, Pentti and Sommer, Friedrich T and Olshausen, Bruno A},
  journal={Neural Computation},
  volume={37},
  number={1},
  pages={1--37},
  year={2024},
  publisher={MIT Press 255 Main Street, 9th Floor, Cambridge, Massachusetts 02142, USA~…}
}

@article{tancik2020fourier,
  title={Fourier features let networks learn high frequency functions in low dimensional domains},
  author={Tancik, Matthew and Srinivasan, Pratul and Mildenhall, Ben and Fridovich-Keil, Sara and Raghavan, Nithin and Singhal, Utkarsh and Ramamoorthi, Ravi and Barron, Jonathan and Ng, Ren},
  journal={Advances in neural information processing systems},
  volume={33},
  pages={7537--7547},
  year={2020}
}

@inproceedings{sitzmann2020siren, publisher = {Curran Associates, Inc.}, booktitle = {Advances in Neural Information Processing Systems (NeurIPS)}, author = {Vincent Sitzmann and Julien N. P. Martel and Alexander W. Bergman and David B. Lindell and Gordon Wetzstein}, title = {Implicit Neural Representations with Periodic Activation Functions}, year = {2020}, url = {http://arxiv.org/abs/2006.09661v1}, entrytype = {inproceedings}, id = {sitzmann2020siren} }

@inproceedings{satorras2021n,
  title={E (n) equivariant graph neural networks},
  author={Satorras, V{\i}ctor Garcia and Hoogeboom, Emiel and Welling, Max},
  booktitle={International conference on machine learning},
  pages={9323--9332},
  year={2021},
  organization={PMLR}
}

@article{strogatz1993coupled,
  title={Coupled oscillators and biological synchronization},
  author={Strogatz, Steven H and Stewart, Ian},
  journal={Scientific american},
  volume={269},
  number={6},
  pages={102--109},
  year={1993},
  publisher={JSTOR}
}

@misc{rw2019timm,
  author = {Ross Wightman},
  title = {PyTorch Image Models},
  year = {2019},
  publisher = {GitHub},
  journal = {GitHub repository},
  doi = {10.5281/zenodo.4414861},
  howpublished = {\url{https://github.com/rwightman/pytorch-image-models}}
}

@inproceedings{keller2023neural,
  title={Neural wave machines: learning spatiotemporally structured representations with locally coupled oscillatory recurrent neural networks},
  author={Keller, T Anderson and Welling, Max},
  booktitle={International Conference on Machine Learning},
  pages={16168--16189},
  year={2023},
  organization={PMLR}
}

@article{DBLP:journals/corr/GehringAGYD17,
  author       = {Jonas Gehring and
                  Michael Auli and
                  David Grangier and
                  Denis Yarats and
                  Yann N. Dauphin},
  title        = {Convolutional Sequence to Sequence Learning},
  journal      = {CoRR},
  volume       = {abs/1705.03122},
  year         = {2017},
  url          = {http://arxiv.org/abs/1705.03122},
  eprinttype    = {arXiv},
  eprint       = {1705.03122},
  bibsource    = {dblp computer science bibliography, https://dblp.org}
}

@article{keller2021topographic,
  title={Topographic vaes learn equivariant capsules},
  author={Keller, T Anderson and Welling, Max},
  journal={Advances in Neural Information Processing Systems},
  volume={34},
  pages={28585--28597},
  year={2021}
}

@article{lowe2023rotating,
  title={Rotating features for object discovery},
  author={L{\"o}we, Sindy and Lippe, Phillip and Locatello, Francesco and Welling, Max},
  journal={Advances in Neural Information Processing Systems},
  volume={36},
  pages={59606--59635},
  year={2023}
}

@article{lowe2022complex,
  title={Complex-valued autoencoders for object discovery},
  author={L{\"o}we, Sindy and Lippe, Phillip and Rudolph, Maja and Welling, Max},
  journal={arXiv preprint arXiv:2204.02075},
  year={2022}
}

@article{miyato2024artificial,
  title={Artificial Kuramoto Oscillatory Neurons},
  author={Miyato, Takeru and L{\"o}we, Sindy and Geiger, Andreas and Welling, Max},
  journal={arXiv preprint arXiv:2410.13821},
  year={2024}
}

@article{NF_Review,
    journal = {Computer Graphics Forum},
    title = {Neural Fields in Visual Computing and Beyond},
    author = {Xie, Yiheng and Takikawa, Towaki and Saito, Shunsuke and Litany, Or and Yan, Shiqin and Khan, Numair and Tombari, Federico and Tompkin, James and Sitzmann, Vincent and Sridhar, Srinath},
    year = {2022},
    publisher = {The Eurographics Association and John Wiley & Sons Ltd.},
    ISSN = {1467-8659},
    DOI = {10.1111/cgf.14505}
}

@article{chi2022kerple,
  title={Kerple: Kernelized relative positional embedding for length extrapolation},
  author={Chi, Ta-Chung and Fan, Ting-Han and Ramadge, Peter J and Rudnicky, Alexander},
  journal={Advances in Neural Information Processing Systems},
  volume={35},
  pages={8386--8399},
  year={2022}
}

@article{sorscher2023unified,
  title={A unified theory for the computational and mechanistic origins of grid cells},
  author={Sorscher, Ben and Mel, Gabriel C and Ocko, Samuel A and Giocomo, Lisa M and Ganguli, Surya},
  journal={Neuron},
  volume={111},
  number={1},
  pages={121--137},
  year={2023},
  publisher={Elsevier}
}

@article{kanerva2022hyperdimensional,
  title={Hyperdimensional computing: An algebra for computing with vectors},
  author={Kanerva, Pentti},
  journal={Advances in Semiconductor Technologies: Selected Topics Beyond Conventional CMOS},
  pages={25--42},
  year={2022},
  publisher={Wiley Online Library}
}
\appendix
\newpage

\section{Broader Impact}

This work is fundamental research. While this work could lead to the discovery of better positional encodings and higher performing visual foundation models, the positivity or negativity of this impact is determined by the downstream task and not this work. 

\section{Limitations}
While our results do not show relative embeddings to be detrimental, we believe them to be evidence that equivariance is not the reason for RoPE's success. 
However, our experiments were performed in Vision where the number of tokens is limited compared to the long context lengths of NLP. 
Moreover, the datasets are not what many believe to be ``at scale". 
While Spherical RoPE and LieRE would intuitively be favored at scale over Axial RoPE, as they have less inductive bias, it is unclear whether inductive bias and equivariance is favored at scale \cite{brehmer2024does}.

It has also been shown that vision is \textit{not} a purely equivariant task and benefits from relaxed equivariance \cite{elsayed2020revisiting}. Our results do not show that equivariance is not useful in tasks that are grounded in physics and obey strict symmetries.

\section{Literature Review}\label{app:litrev}

\subsection{Natural Language Processing}

In natural language, positional encoding has been used to break the permutation, ``bag of words", symmetry \cite{vaswani2017attention}. Although this could be done by learning a vector per position, this is both memory-expensive  for large context sizes making it practical to apply to only the first layer. Moreover, it does not allow for extrapolation at test time to context sizes beyond training. Thus, it is favorable to perform positional embeddings with a predictable deterministic function. One way of doing this is to make the attention relative with local receptive fields, as is done implicitly in convolutional neural networks \cite{chi2022dissecting}. Sinusoidal positional embeddings were proposed due to approximate local and shift-invariant properties of Random Fourier Features \cite{rahimi2007random}. Since sinusoidal, other methods have been proposed to get guaranteed shift invariance by explicitly parameterizing based on distance \cite{shaw2018self, pham2020relative, press2021train}. However, these methods require a positional embedding for every pair of positions which is not supported by many of the efficient attention optimizations such as Flash attention \cite{dao2022flashattention} \cite{angelotti2023hype}.

Rotary Positional Embeddings (RoPE) have become the staple in NLP having recently been adopted by many of the large language models \cite{wang2024qwen2,grattafiori2024llama, team2024gemma, liu2024deepseek, jiang2024mixtral}. However, these methods also use causal masking, which has been shown to allow models with no positional embedding to recover absolute position \cite{haviv2022transformer, yang2025rope, wang2024length,kazemnejad2023impact}. This has led to questions on the importance of relative position \cite{barbero2024round}.

In language, there has also been extensions to RoPE proposed through NTKs and kernel methods \cite{chi2022kerple}. However, these methods have not, to our knowledge, seen use in vision.

\subsection{Vision and Video}
Vision transformers were introduced in \citet{dosovitskiy2020image} and, though they tried sinusoidal position encodings, found learnable position encodings to perform best. For convolution-esque models such as SWin transformers, relative positional encodings have been popular \cite{liu2021swin,chu2021conditional}. More recently, RoPE has been shown to be an efficient and simple way to have relative embeddings and has been extended to 2D using Axial and Mixed RoPE. Going beyond 2D to Video data, Axial RoPE has become increasingly popular. The extension was first attributed to \citet{wang2024qwen2} as 3D-RoPE or M-RoPE, leading to two separate Video-RoPE papers from \citet{wei2025videorope} and \citet{liu2025vrope}. Both of these focus on the order of the position enumeration and interleaving positions. However, this should not be a problem if frequencies are not deterministic, \textit{or} if frequencies are indexed by both $d$ and modality $m$ as done in Eq \ref{eq:modality}. We highly recommend using either Mixed RoPE or LieRE which extend naturally for videos.

LieRE embeddings have thus far been the most general form of RoPE to $N$-D. However, \citet{schenck2025learning} has claimed the method to have a large memory footprint and proposed STRING. This paper, a preprint released concurrently with the writing of this manuscript, follows much of the same math as this paper. However, they did not recognize that an orthogonal matrix is implicitly learned by the query and key matrix. Moreover, their method relies on commuting Lie algebras. From our insights in Section \ref{sec:generality}, their method can likely be viewed as a slower implementation of $N$-D Mixed-RoPE.

It is also worth noting that positional encodings have also been explored within vision through the area of Neural Fields \cite{NF_Review}. Traditional coordinate MLPs have been found to be biased toward low-frequency functions \cite{tancik2020fourier} leading to more advanced positional encodings such as Random Fourier Features \cite{rahimi2007random} or sinusoidal activation functions \cite{sitzmann2020siren}. These implicit functions have been used to encode attention and message passing in graph neural networks with recent work being put in to make these functions equivariant to symmetry transformations \cite{satorras2021n, brandstetter2021geometric, knigge2024space}.

\subsection{Graphs and AI in Science}

Positional encodings are well studied within graph neural networks \cite{li2020distance, park2022grpe}. Graphs are limited in their expressivity up to the  Weisfeiler-Lehman (WL) graph isomorphism test \cite{xu2018powerful}, so positional encodings can break the isomorphism symmetry \cite{he2023sheaf,you2019position}. Within this community, they propose \textit{spectral attention} and graph Laplacians for positional encoding \cite{kreuzer2021rethinking}. These methods seem extremely close to our analysis of RoPE, but from a very different perspective. We show that the frequencies of RoPE can be interpreted as the eigenvalues of an orthogonal transformation by taking the spectral decomposition.

In an overlapping vein, relative position encodings have been studied in terms of equivariant graph neural networks, often for scientific disciplines such as molecular physics \cite{brandstetter2021geometric, schutt2021equivariant} or drug discovery \cite{jumper2021highly}. One method to achieve equivariance is through defining relative coordinate frames \cite{kofinas2021roto}. This corresponds to the learned relative positional method described in \citet{shaw2018self}, but can be generalized to higher dimensions and different transformation using bi-invariant distance functions \cite{bekkers2023fast,knigge2024space,wessels2024grounding}. The message-passing functions of these works correspond to a generalization of attention scores \cite{gilmer2017neural}.

However, even in these tasks with physics-grounded symmetries, the need for equivariance is hotly debated. While AlphaFold \cite{jumper2021highly} was originally touted as the example of the success of equivariant inductive biases in science, AlphaFold 3 \cite{abramson2024accurate} explicitly stated that they benefited from removing this inductive bias at scale. However, while the harm of inductive bias at scale is the prevalent zeitgeist, it is not an established fact \cite{brehmer2024does}.

\subsection{Computational Neuroscience}

Coupled oscillators have become a growing area of interest within computational neuroscience \cite{keller2021topographic,keller2023neural, strogatz1993coupled}. By observing the projection of the RoPE circles onto the real axis, one can interpret RoPE as time progression in $D$ uncoupled, undamped harmonic oscillators. This perspective naturally connects RoPE to \citet{lowe2022complex}'s series of papers on complex autoencoders and their extensions \cite{lowe2023rotating,miyato2024artificial}.

In another, vein of research, there has been some work in hyper-dimensional computing\cite{kanerva2009hyperdimensional, kanerva2022hyperdimensional} in Phasor and Residue VSAs \cite{kymn2024computing} which represent concepts as rotations around unit circles in high-dimensional spaces. These representations have strong connections with RoPE. Additionally, progress has been made in hypothesizing how biological neural networks encode positional knowledge with hexagonal grid cells, which can be represented as a discrete sum of three periodic functions oriented at the cubic roots of unity\cite{sorscher2023unified}.

\subsection{Generality of RoPE}

The generality of RoPE has been found by others. \citet{schenck2025learning}, \citet{NDRoPE}, and \citet{liu2025rethinking} all propose proofs similar to Proposition \ref{th:generality}. However, \citet{schenck2025learning} miss that the orthogonal transformation can be incorporated into key matrix. \citet{liu2025rethinking} and \citet{NDRoPE} take the assumption of \textit{reversibility}, which leads to the independent eigenvalue assumptions of Axial RoPE. All three works take the assumption of an abelian subgroup -- i.\,e. commutative generators, -- but miss the generality of Mixed RoPE. While \citet{NDRoPE} propose quaternions -- i.\,e. spherical rotations -- as a direction, they immediately dismiss it as a \textit{no-go} because they lack equivariance. This exemplifies the ``circular argument," where equivariance is assumed to be necessary because work will not investigate non-equivariant positional encodings because equivariance is necessary.

Because our derivation was found independently of these works and the previous works are, to our knowledge, not published, we have left in Proposition~\ref{th:generality}. We would like to acknowledge their work, but retain the flow of this paper.

\section{Notation}
\label{app:notations}
\begin{table}[h]
\centering
\resizebox{\linewidth}{!}{
\begin{tabular}{|c|c|c|p{6cm}|}
\hline
\textbf{Symbol / Term} & \textbf{Dimension} & \textbf{Meaning} & \textbf{Notes} \\ \hline
$\mathbf{x}_i$ &$\mathbb{R}^D$ & Patch/token/content vector of token $i$ & Raw input embedding \\ \hline
$x_i$ & $\mathcal{X}$ & Abstract content of token $i$ & Raw input embedding \\ \hline
$p_i$ & $\mathbb{R}^M$ or $\mathcal{P}$ & Position of token $i$, can be $M$-D or abstract $\mathcal{P}$ & Scalar (1D) or vector (2D) \\ \hline
$m$ & $\mathbb{Z}$& Modality index & e.g., $x$, $y$, time \\ \hline
$M$ & $\mathbb{Z}$& Number, or space, of Modalities & \\ \hline
$D$ & $\mathbb{Z}$& Hidden dimension & Number of pairs/triples/quadruples \\ \hline
$T$ & $\mathbb{Z}$& Number of Tokens & \\ \hline
$\mathbf{W}_q, \mathbf{W}_k, \mathbf{W}_v$ & $\mathbb{R}^{\mathcal{X} \times D}$& Query, Key, Value Matrices & \\ \hline
$\mathbf{q}$ & $\mathbb{R}^{N}$ & $\mathbf{q}_i = \mathbf{W}_q x_i$& Query vector \\ \hline
$\mathbf{k}$ & $\mathbb{R}^{N}$& $\mathbf{k}_j = \mathbf{W}_k x_j$& Key vector \\ \hline
$\mathbf{v}$ & $\mathbb{R}^{N}$&$\mathbf{v}_j = \mathbf{W}_v x_j$& Value vector \\ \hline
$\mathbf{Q}, \mathbf{K}, \mathbf{V}$  & $\mathbb{R}^{T \times N}$ & Query, Key, Values & $T$ tokens, $D$ latent dimensions \\ \hline
$\varphi(x, p)$ & $\mathcal{X} \times \mathcal{P} \to \mathbb{R}^D$& Positional Encoding function &  \\ \hline

$\mathbf{Z}$ & $\mathbb{R}^{T \times N}$& Output of Attention & $\mathbf{Z} = \text{Attention}(\mathbf{Q}, \mathbf{K}, \mathbf{V})$ \\ \hline
$a(i, j)$ & $\mathbb{R}$ & Attention weight & Softmax of attention scores \\ \hline
$\alpha(\mathbf{q}, \mathbf{k})$& $\mathbb{R}$ & Attention score & Inner product $\mathbf{q}^\top \mathbf{k}$ \\ \hline
$\omega_d$/$\lambda_d$ &$\mathbb{R}$& Rotation frequency for dimension $d$ & Equivalent to eigenvalue of generator \\ \hline
$\mathbf{q}_d$ & $\mathbb{R}^{2/3/4}$ & Query pair/triple/quadruple at dimension $d$ & After RoPE or LieRE applied \\ \hline
$\mathbf{R}_{\omega_d p}$ & $\mathbb{R}^{2\times 2}$&$2 \times 2$ rotation matrix & Rotation based on frequency and position \\ \hline
\end{tabular}
}
\caption{Summary of Notations and Key Concepts}
\label{tab:notations}
\end{table}

\newpage

\begin{table}[h]
\centering
\resizebox{\linewidth}{!}{
\begin{tabular}{lccccccl}
\toprule
\textbf{Positional Encoding} & \textbf{Vision} & \textbf{Learned} & \textbf{Extrapolation} & \textbf{QK Separable} & \textbf{Relative} & \textbf{Linear Flow} & \textbf{Used In} \\
\midrule
Absolute (Sinusoidal) & \xmark & \cmark/\xmark & \cmark & \cmark & \cmark & \xmark &  Transformer\cite{vaswani2017attention} \\
Absolute (Learned)    & \cmark & \cmark & \xmark & \cmark & \cmark & \xmark & BERT, GPT, ViT\cite{dosovitskiy2020image} \\
Absolute (Random-Fourier) & \xmark  & \xmark & \cmark & \cmark & \xmark & \cmark & FNet\cite{lee2021fnet}, Performer \cite{choromanski2020rethinking} \\
Relative (Learned)    & \xmark & \cmark & \xmark & \xmark & \xmark & \xmark& Transformer-XL, T5 \cite{raffel2020exploring} \\
ALiBi                 & \xmark & \cmark/\xmark & \cmark & \cmark & \cmark & \cmark & LLaMA 2 \cite{grattafiori2024llama}, ALiBi \cite{press2021train} \\
NoPE                 & \xmark* & \xmark & \cmark* & \cmark* & \cmark* & \cmark* & LLaMA 4 \cite{meta_llama4_2025}
\\
Rotary (RoPE)         & \xmark & \xmark & \cmark & \cmark & \cmark & \cmark & Contemporary LLMs \cite{wang2024qwen2,grattafiori2024llama,team2024gemma,jiang2024mixtral} \\
\bottomrule

Axial RoPE         & \cmark & \cmark/\xmark & \cmark & \cmark & \cmark & \cmark & VisionLLaMA\cite{chu2024visionllama}, Qwen2\cite{wang2024qwen2}, VideoRoPE\cite{wei2025videorope}\\
Mixed RoPE         & \cmark & \cmark & \cmark & \cmark & \cmark & \cmark & \citet{heo2024rotary}\\
LieRE         & \cmark & \cmark & \cmark & \cmark & \xmark & \cmark & \cite{ostmeier2024liere}\\
\bottomrule
Spherical RoPE         & \cmark & \cmark/\xmark & \cmark & \cmark & \xmark & \cmark & Ours \\
Uniform RoPE         & \cmark & \cmark/\xmark & \cmark & \cmark & \cmark & \cmark & Ours \\

\bottomrule
\end{tabular}
}
\caption{Comparison of positional encoding methods in transformer models. $^*$NoPE makes some properties trivially true.}
\end{table}

\section{Positional Encoding Properties}\label{app:properties}

Rotary positional embeddings were derived in \citet{su2024roformer} by drawing equations from assumed properties. While these appear as arithmetic assumptions and equations in their work, we formalize what properties these assumptions imply and why we may choose these assumptions in this section. In their paper, to derive their equations, they use equivariance (relativity), query-key separability of the positional encoding, linearity and incompressability, locality, and query-key symmetry.
\begin{enumerate}
    \item Equivariance/Relativity: Attention score should be affected only by the relative position of two tokens, i.\,e. have the form
    \begin{equation}
        \alpha(x_i,x_j,p_i,p_j) = \hat\alpha (x_i,x_j,p_i - p_j).
    \end{equation}    
\item Key-query seperability: The positional encoding, $\varphi$, of the query should not depend on the position of the key
    \begin{equation}
        \alpha(x_i,x_j,p_i,p_j) = \bar\alpha (\varphi(x_i,p_i),\varphi(x_j,p_j)) 
    \end{equation}
    \item Linearity: The positional encoding should be a linear flow, see Appendix \ref{app:linearflow}. Namely,
        \begin{equation}
        \varphi(\varphi(x,p_i), p_j) = \varphi(x,p_i+p_j).
        \end{equation}
    \item Locality: The attention score between two tokens should decay with distance 
        \begin{equation}
            \lim_{|p_i-p_j|\to\infty}\alpha(x_i,x_j,p_i,p_j)=0
        \end{equation}
\end{enumerate} 
\subsection{Relativity and Equivariant}\label{app:equiv}

We use the term \textit{equivariant} interchangably with \textit{relative}. Strictly speaking, one should specify the transformation or group you would like to be relative to, e.\,g. shift/rotation or $SO(2)$. As previous literature always refers to relative positional bias in terms of shifts/translations, in the main text, this is what we mean. We use the term equivariance to be the generalization of relativity beyond language because we would like to refrain from using the term ''relativity" to describe the property of being a relative PE too often due to its connotation within theoretical physics. First, we define relative in the case of positional encodings in language as
\begin{equation} \label{eq:rel}
    \alpha(x_i,x_j, p_i, p_j) = \hat{\alpha}(x_i,x_j, p_i-p_j).
\end{equation}
In the rest of this section, we mathematically explore where this equation comes from.

The behavior we are trying to capture is that if we renumber the words in the sentence, it should not affect the attentions score. Intuitively, if a text is padded with spaces at the beginning, that will not have a significant effect on the meaning of the sentences. We can ensure this by colloquially saying that the attention between two words should depend on the distance between them. Notice, that strictly speaking this is not a proper distance, since it can be negative; it is, instead, a \textit{signed} distance function. Though this may seem pedantic in one dimension, in two dimensions defining a distance function is less unique. For example, one may choose $\mathbb{L}_1$ or $\mathbb{L}_2$ distance metrics. 
Because distance functions are more nebulous, it makes more sense to define relative in terms of the transformations that we would like our attention score to be independent of.
\begin{equation}
    \alpha(x_i,x_j, p_i, p_j) = \alpha(x_i,x_j, T(p_i), T(p_j)).
\end{equation}
These transformations can be combined to generate a set of transformations which leave the attention score unchanged, or \textit{symmetric}. This set has the mathematical properties of a group and is known as a symmetry group. We can index transformations by elements in the symmetry group, $g\in G$, and let the elements act on 
\begin{equation}
    \alpha(x_i,x_j, p_i, p_j) = \alpha(x_i,x_j, g.p_i, g.p_j).
\end{equation}
As an example, $g$ could represent an angle, $\theta$, and it may act on a vector $\mathbf{p}$ as a rotation $g.\mathbf{p} = \mathbf{R}_\theta\mathbf{p}$.

Connecting everything back to Eq. \ref{eq:rel}, Noether's theorem states that any continuous symmetry can be expressed as a conservation law. This allows us to introduce bi-invariant function \cite{knigge2024space, wessels2024grounding}, or ``Noether charge", $\beta(p_i,p_j)$, that is invariant under the group action,
\begin{equation}
    \beta(p_i, p_j) = \beta(g.p_i, g.p_j) \implies \beta(p_i, p_j) - \beta(g.p_i, g.p_j) =0.
\end{equation}
Thus, we can express our symmetry group through isodistances of $\beta$, 
\begin{equation}
    \alpha(x_i,x_j p_i,p_j) := \hat{\alpha}(x_i,x_j,\beta(p_i,p_j)).
\end{equation}
For example, we can pick the function
\begin{equation}
    \beta(p_i,p_j)  = p_i-p_j = (p_i-p_0) - (p_j - p_0) = \beta(p_i-o,p_j-p_0)
\end{equation}
If we were to define $\beta(p_i,p_j) = |p_i-p_j|$, then we we would additionally be equivariant to reflection of the order of tokens in a sentence. If we trivially define $\beta(p_i,p_j)= C$, then we arrive at bag of words, or no positional encoding (NoPE). For a list of common transformations and their corresponding bi-invariants see Theorem 1 of \citet{bekkers2023fast}.

\subsection{Query-Key Separability}

Query and key separability is important for efficiency reasons. If we can decompose our positional encoded attention score as,
\begin{equation}
    \alpha(x_i,x_j,p_i, p_j) = \alpha(\varphi(x_i,p_i), \varphi(x_j,p_j))
\end{equation}
then we can pre-compute the positional encoding for the queries and keys on time making the computation $O(T)$. If the positional encoding is not separable, then it will need to be computed for \textit{every pair}, $(i,j)$\cite{liu2021swin,raffel2020exploring, shaw2018self}. Although there are many symmetries that can be exploited to make this not a quadratic computation, it removes the symmetries exploited by efficient attention mechanisms \cite{rope-eleutherai, choromanski2020rethinking, katharopoulos2020transformers}.

\subsection{Linear Flow Property}\label{app:linearflow}

The property of being a ``flow" was first proposed in \citet{DBLP:journals/corr/abs-2003-09229}, however it is not often discussed. It is a property inherently present in RoPE\cite{su2024roformer}, LieRE\cite{ostmeier2024liere} and ALiBi \cite{press2021train} embeddings, specifically as a \textit{linear flow}.

We use the term \textit{linear flow} for this property because the embedding can be found by repeated application of a linear function. However, the term ``linear" this is a small misnomer because it is only \textit{locally} linear. We define a \textit{flow} as function
\begin{equation}
    \varphi : \mathbb{R}^N\times \mathbb{R} \to \mathbb{R}^N
\end{equation}
such that for all $x\in X$ and $p_1,p_2\in \mathbb{R}$, the following conditions hold:
\begin{enumerate}
    \item Initial condition (identity at time zero):
    \begin{equation}
        \varphi(0,x) = x
    \end{equation}
    \item Group property (flow property):
    \begin{equation}\label{appeq:flow}
        \varphi(\varphi(\mathbf{x}, p_1), p_2)) = \varphi(x,p_1 +p_2)
    \end{equation}
    \item Continuity (or differentiability): $\varphi$ is continuous with respect to its variables, depending on the context
\end{enumerate}
Strictly speaking, continuity is not necessary for positional encodings as positions tend to be integer values. What we really wish to capture with this property is for the positional encoding to be recursively defined. It may be strange to wish to apply the positional encoding multiple times; however, by having the positional encoding as an endomorphism it can allow for more predictable behavior when extrapolating to larger contexts, which we suspect helps the model train.

We define a position embedding to be a \textit{linear flow} if the flow has the form:
\begin{equation}
    \varphi(\mathbf{x},\Delta p) = \mathbf{A}\mathbf{x},
\end{equation}
for $\mathbf{A} \in \mathbb{R}^{N\times N}$ and $\mathbf{x}\in\mathbb{R}^N$, where $\Delta p$ is the increment rate for position. By Eq. 
\ref{appeq:flow}, any position $p := p_0 \Delta p$ can then be attained by,
\begin{equation}
    \varphi(\mathbf{x}, p) = \mathbf{A}^{p_0}\mathbf{x}.
\end{equation}
This can be seen as a \textit{geometric series} if $\mathbf{A}$ is a scalar as seen in \citet{press2021train}. If we let $\Delta t$ become infinitesimal, then we can express the recurrence relationship as the ODE,
 \begin{equation}
     \frac{\partial\varphi}{\partial t} = \mathcal{A}\varphi
 \end{equation}
which we can integrate to get,
\begin{equation}
    \varphi(\mathbf{x}, p) = \exp({\mathcal{A}p})\mathbf{x}
\end{equation}
This $\mathcal{A}$ is our \textit{generator} of the flow, which is also a generator for a \textit{matrix Lie algebra}, which we focus on in the main text. The matrix exponential, $\exp : \mathbb{R}^{N\times N} \to \mathbb{R}^{N\times N}$, can be unstable for long contexts; similar to the scalar exponential function $e^{xp}$, the function can quickly become large for high values of $x$. However, this can be stable value $x=0$, since it always results in one. Similarly, the matrix exponential can be stable if the divergence of the flow -- trace of the generator -- is zero. 
We call flow ``incompressible" or ``divergence-free" if the trace of $\mathcal{A}$ is zero, making the determinant of $\mathbf{A}$ unit. If fluid dynamics, this is called \textit{incompressibility}. For fluids, this implies that the flow conserves mass.

If there are more than one generator of the Lie group, $\mathcal{A}_1$ and $\mathcal{A}_2$, then Eq.~\ref{appeq:flow} must be modified to,
\begin{equation}
        \varphi(\varphi(\mathbf{x}, \mathbf{p_1}), \mathbf{p_2})) = \varphi(\mathbf{x},\mathbf{p}_1 \circ \mathbf{p}_2),
    \end{equation}
where $\circ$ is the group product.
By the Baker–Campbell–Hausdorff formula, $\exp{\mathcal{A}_1p_1}\exp{\mathcal{A}_2p_2} = \exp{\mathcal{A}_1p_1 + \mathcal{A}_2p_2}$ iff the commutator of $\mathcal{A}_1p_1$ and $\mathcal{A}_2p_2$ is zero, i.\,e. the matrices commute. If they do commute, then 
\begin{equation}
        \varphi(\varphi(\mathbf{x}, \mathbf{p_1}), \mathbf{p_2})) = \varphi(\varphi(\mathbf{x}, \mathbf{p_2}), \mathbf{p_1}))  \implies \varphi(\mathbf{x}, \mathbf{p}_1 \circ \mathbf{p}_2) = \varphi(\mathbf{x}, \mathbf{p}_2 \circ \mathbf{p}_1)
\end{equation}
thus making $\circ$ commutative and having the same properties as addition, $\circ$ := ``$+$", and Eq.~$\ref{appeq:flow}$ will hold. In this case, the group/flow is known as an \textit{abelian} Lie group, or \textit{abelian flow}. However, if they do not commute, then $\circ$ will not commute and they are known as \textit{non-abelian}. This also makes the flow \textit{non-integrable}.


\subsection{Locality}
Locality is often conflated with relativity. The general idea is that tokens far from each other should be independent of one another -- i.\,e. attention should decay as distance grows. This often motivates the definition
\begin{equation}
    \lim_{|p_i-p_j| \to \infty} \alpha(x_i,x_j,p_i,p_j) = 0
\end{equation}
for $p_i,p_j\in \mathbb{R}$ and $x_i,x_j\in\mathbb{R}^D$.
However, this definition is \textit{both} relative and local. We instead define local as,
\begin{equation}
    \lim_{|p_i-p_0| \to \infty} \alpha(x_i,x_j,p_i,p_0) = 0.
\end{equation}
The difference being that $p_0$ is the \textit{origin} position. If an embedding is relative, then the origin is arbitrary and can be defined as $p_i$ or $p_j$. In \citet{press2021train}, they define the origin vector as the next word. However, they can only do this because of the causal mask. 

In general, the most natural way to measure locality is through the concept of the quantum mechanical concept of the \textit{variance of an operator}. We will simply use exponential decay, but we point interested readers to Chapter 3 of \citet{Griffiths}. This formalism works for RoPE as it is a linear transformation and the attention mechanism defines a Hilbert space. 

To be clear, RoPE and LieRE are \textit{not} local embeddings. This was shown for RoPE in \citet{barbero2024round}. Because they are orthogonal matrices, they have unit determinant, which naturally precludes locality. 

\subsection{Other properties}

For completeness, there are two additional assumptions that are common. 
\paragraph{Adjoint symmetry of the Positional Encoding}
We implicitly assume that the positional encoding is symmetric for the query and key. That is, we assume that the query and key are from the same domain, so the positional encoding has the same representation. More generally, the positional encoding can act differently on the query and key,
\begin{equation}
    \alpha( \bar{\varphi}(x_i,p_i), \varphi(x_j, p_j)) = \alpha( \varphi(x_i,p_i), \varphi(x_j, p_j)),
\end{equation}
where $\bar{\varphi}$ is the positional encoding function for queries.
More generally, we can have a relative embedding by letting $\bar{\varphi}$ act on queries differently from the keys. For example, if we let 
\begin{align}
    \varphi(x,p) &= \exp(\Lambda p) & \bar{\varphi}(x,p) &= \exp(-\Lambda p),
\end{align}
where $\Lambda$ is a diagonal matrix. We end up with,
    \begin{equation}
    \alpha( \bar{\varphi}(x_i,p_i), \varphi(x_j, p_j)) = \mathbf{q}_i^\top \exp(\Lambda(p_j-p_i)) \mathbf{k}_j,
\end{equation}
where RoPE can be interpreted as a simple harmonic oscillator, by weakening the symmetry requirement, one could incorporate damping. This can also be used to incorporate graph Laplacian positional encodings into the framework.

\paragraph{Reversibility}

Reversibility means that the positional encoding is an injective map -- that is, every coordinate is mapped to a unique rotation, thus position can be recovered. This property is important in \citet{liu2025rethinking} and \citet{NDRoPE} to derive Axial RoPE. While it prevents Eq. \ref{naive_rope}, it is necessary only for the $D=1$ case. More generally, Mixed RoPE can learn an injective map for large $D$. Moreover, while having a ``lossless" positional encoding is nice mathematically, its practical utility has yet to be soundly justified, especially if the positional encoding is learnable.

\newpage
\section{Fast Implementation} \label{app:Fast Implementation}
We follow a vectorized implementation for Spherical RoPE similar to the ``fast implementation" proposed in \citet{su2024roformer}.

First, apply the rotation directly on after the other:
\begin{align}
    z_d[1] &= \cos(\omega_y p_y)~z_d[1] - \sin(\omega_y p_y)~z_d[3]\label{1}\\
    z_d[3] &= \sin(\omega_y y)~z_d[1] + \cos(\omega_y)~z_d[3],\label{2}
\end{align}
then
\begin{align}
    z_d[2] &= \cos(\omega_y p_x)~z_d[2] - \sin(\omega_x p_x)~z_i[3]\label{3}\\
    z_d[3] &= \sin(\omega_x p_x)~z_d[2]  + \cos(\omega_xp_x)~
    z_d[3]\label{4},
\end{align}
where steps \ref{1} and \ref{2} happen simultaneously, and steps \ref{3} and \ref{4} occur at the same time.

\section{Experimental Setup}

\paragraph{Models} We use the ViT-S backbone from the timm library \cite{rw2019timm}. The network always has a depth of 12. We keep $N$ as close to constant across models as we can. For CIFAR100, the embedding dimensions are changed from $64\times N_\text{heads}$ to $60\times N_\text{heads}$ to be compatible with pairs, triplets and quadruples. For ImageNet, we make the embedding dimension $63\times N_\text{heads}$ for Spherical RoPE and $64\times N_\text{heads}$ for other methods. For classification, we use a class token to pool the tokens and predict. Unlike the patch tokens, the class token is not affected by any positional encoding.

\paragraph{CIFAR100}
All experiments on CIFAR100 were performed on one A100 GPUs with a batch size 256. We use a patch size of $4\times 4$ on the original image size $32 \times 32$. The training uses heavy regularization and augmentations including dropout, MixUp \cite{zhang2017mixup} and CutMix \cite{yun2019cutmix}. The models are trained for 400 epochs, taking $\sim$ 40 seconds per training loop.

\paragraph{ImageNet}
All experiments on ImageNet-1k were performed on four A100 GPUs with a batch size 256. We used cosine learning rate with a learning rate of $3e-3$ for 200 epochs with 5 epochs of linear warm-up.  We used a patch size of $16\times 16$ on the cropped and resized $224 \times 224$ image after applying 3-Augment \cite{touvron2022deit}. We use the LAMB \cite{you2019large} optimizer. All experiments took $\sim$20 hrs with $\sim 5$ to $8$ minutes to complete a training loop depending on method.

\paragraph{Positional Encodings} For testing with different resolutions, the images from ImageNet's validation set were normalized, resized and cropped. On training, the patches were assigned position $[-\pi, \pi]$ and for evaluation, the patch positions were extrapolated to the range $[-\frac{P}{P_0} \pi, \frac{P}{P_0} \pi]$. For Learned APE, the positional embeddings are instead interpolated. The fixed frequencies were given by $\omega_d = 1/100^{2d/D}$, where $d$ is the index of the pair/tuple/quadruple. One frequency is shared between both $x$ and $y$ in our implementation of Axial RoPE
. 

\newpage

\section{Hyperparameters}
\label{app:hyperparameters}
\begin{table}[ht]
\centering
\caption{Hyperparameters for ImageNet-1K Training}
\begin{tabular}{@{}ll@{}}
\toprule
\textbf{Category} & \textbf{Setting} \\ \midrule

\multicolumn{2}{@{}l}{\textbf{Model Architecture}} \\
Patch Size                  & 16x16 \\
Heads                       & 6 \\
Latent Dimension            & 64 (63 for Spherical) $\times$ Heads \\
Depth                       & 12 \\
Pooling                     & [CLS] \\
Stochastic Depth            & No \\
Dropout                     & No \\
LayerScale                  & 1 \\

\addlinespace
\multicolumn{2}{@{}l}{\textbf{Optimization}} \\
Optimizer                   & LAMB \cite{you2019large} \\
Base Learning Rate          & 4e-3 \\
Weight Decay                & 0.05 \\
Learning Rate Schedule      & Cosine Decay \\
Warmup Schedule             & Linear \\
Warmup Epochs               & 5 \\
Epochs                      & 200 \\
Batch Size                  & 512 \\
Gradient Clipping           & \cmark \\

\addlinespace
\multicolumn{2}{@{}l}{\textbf{Precision and Backend}} \\
Precision                   & Mixed (bfloat16) \\
Backend                     & torch.autocast \\

\addlinespace
\multicolumn{2}{@{}l}{\textbf{Data Augmentation - Train}} \\
Crop                        & RandomResizedCrop (192$\rightarrow$224) \\
Flip                        & \cmark \\
3-Augment                        & \cmark \\
Color Jitter                & (0.3, 0.3, 0.3, 0.0) \\
Mixup \cite{zhang2017mixup}                       & \xmark \\
Cutmix \cite{yun2019cutmix}                      & \xmark \\ 
Normalization               & ImageNet-1K Statistics \\

\addlinespace
\multicolumn{2}{@{}l}{\textbf{Data Augmentation - Test}} \\
Resize                      & Resize $\rightarrow$ Resolution \\
Crop                        & CenterCrop \\
Normalize                   & ImageNet-1K Statistics \\

\bottomrule
\end{tabular}
\end{table}
\newpage

\begin{table}[ht]
\centering
\caption{Hyperparameters for CIFAR100 Training}
\begin{tabular}{@{}ll@{}}
\toprule
\textbf{Category} & \textbf{Setting} \\ \midrule

\multicolumn{2}{@{}l}{\textbf{Model Architecture}} \\
Patch Size                  & 16x16 \\
Heads                       & 12 \\
Latent Dimension            & 60 $\times$ Heads \\
Depth                       & 12 \\
Pooling                     & [CLS] \\
Stochastic Depth            & 0.1 \\
Dropout                     & 0.1 \\
LayerScale                  & \cmark \\

\addlinespace
\multicolumn{2}{@{}l}{\textbf{Optimization}} \\
Optimizer                   & LAMB \cite{you2019large} \\
Base Learning Rate          & 4e-3 \\
Weight Decay                & 0.05 \\
Learning Rate Schedule      & Cosine Decay \\
Warmup Schedule             & Linear \\
Warmup Epochs               & 5 \\
Epochs                      & 400 \\
Batch Size                  & 1024 \\
Gradient Clipping           & \cmark \\

\addlinespace
\multicolumn{2}{@{}l}{\textbf{Precision and Backend}} \\
Precision                   & Mixed (bfloat16) \\
Backend                     & torch.autocast \\

\addlinespace
\multicolumn{2}{@{}l}{\textbf{Data Augmentation - Train}} \\
Crop                        & RandomResizedCrop (32) \\
Flip                        & \cmark \\
3-Augment                        & \cmark \\
Color Jitter                & (0.3, 0.3, 0.3, 0.0) \\
Mixup \cite{zhang2017mixup}                       &  0.8 \\
Cutmix \cite{yun2019cutmix}                      & 1.0 \\
Normalization               & CIFAR Statistics \\

\addlinespace
\multicolumn{2}{@{}l}{\textbf{Data Augmentation - Test}} \\
Normalize                   & CIFAR Statistics \\

\bottomrule
\end{tabular}
\end{table}
\newpage

\section{Additional Evaluations}\label{app:AddResults}

 In this section, we include extra evaluations including, basic data scaling, segmentation and speed. We also include additional experiments on the effect of rotation frequencies on Uniform RoPE.
\subsection{Data Scaling}

Below we evaluate the data scaling of each method. We partition the CIFAR100 dataset into smaller subsets. The number of epochs is scaled, so that the number of training steps is matched on the data subsets. Each model is trained only once on each data split. This experiment tests whether a commutative constraint is beneficial in smaller data regimes as an inductive bias.

{\captionof{table}{
    Performance on different portions of CIFAR100.
}
\begin{center}
    
\resizebox{.85\linewidth}{!}{

\begin{tabular}{lccccc}
    \toprule
    \textbf{Dataset Size} & \textbf{Spherical (Learned)} & \textbf{Axial (Learned)} & \textbf{Mixed} & \textbf{Uniform} & \textbf{APE} \\
    \midrule
    0.2 & 56.04 ($\mathbf{57.2}$) & 55.3 (56.6) & 56.9 & 52.82 & 45.9 \\
    0.4 & 63.6 ($\mathbf{65.34}$) & 63.3 (62.5) & 64.4 & 59.7 & 53.4 \\
    0.6 & 67.6 (69.8) & 66.0 (66.78) & $\mathbf{70.0}$ & 64.1 & 57.7 \\
    0.8 & 69.8 ($\mathbf{72.6}$) & 69.9 (69.1) & 71.6 & 65.8 & 59.0 \\
    \bottomrule
\end{tabular} 
}
\end{center}
}

Equivariance, in theory, should provide better performance at small scales due to its inductive bias. However, we observe that learned Spherical RoPE performs on-par or better than Mixed RoPE with less parameters. The small gap

\subsection{Segmentation}
Below we include rudimentary experiments on segmentation to show that the equivalent performance of Spherical RoPE is not caused by the simplicity of classification as a task. For these experiments, we use the models trained on ImageNet-1k as pretrained backbones and fine-tune for Pascal VOC Segmentation \cite{Everingham15}. The heads of the models are replaced with a single MLP which is used to get patch logits for each of . Bilinear interpolation is used to create individual pixel logits.
\captionof{table}{
    Segmentation results (IoU) on VOC with and without augmentation.
}
{\begin{center}
\resizebox{\linewidth}{!}{
\begin{tabular}{lccccc}
    \toprule
    & \textbf{Spherical} & \textbf{Axial (Learned)} & \textbf{Mixed} & \textbf{Uniform} \\
    \midrule
    VOC (No Aug.) & $\mathbf{0.45}$($\mathbf{0.46}$) & 0.42 (0.43) & 0.44 & 0.41 & \\
    VOC (Simple Aug.) & $\mathbf{0.498}${\tiny $\pm.007$} ($\mathbf{0.502}${\tiny $\pm.012$}) & 0.474{\tiny $\pm .011$} (0.468{\tiny $\pm .010$}) & $\mathbf{0.502}${\tiny $\pm.008$} & 0.461{\tiny $\pm.012$} & \\
    \bottomrule
\end{tabular}}
\end{center}
}

\subsection{Wall Clock Time}

Below we include the wall clock time for each method. Beyond vectorization as described in Appendix \ref{app:Fast Implementation}, no optimizations were made for speed. LieRE was implemented following the pseudo-code in \citet{ostmeier2024liere}.

\begin{table}[h!]
\centering
\caption{Time comparison across different positional encodings}
\resizebox{\linewidth}{!}{
\begin{tabular}{lcccccc}
\toprule
\textbf{Time comparison} & \textbf{Spherical (Learned)} & \textbf{Axial (Learned)} & \textbf{Mixed} & \textbf{LieRE} & \textbf{APE} & \textbf{Uniform} \\
\midrule
Without \texttt{torch.autocast} & 16.6s (16.6s) & 16.5s (16.7s) & 15.7s & 27.4s & 13.1s & 16.5s \\
With \texttt{torch.autocast} & 6.7s (5.8s) & 6.5s (5.7s) & 5.2s & 13.6s & 3.9s & 6.6s \\
\bottomrule
\end{tabular}}
\end{table}

The experiment was performed by running a dummy input of dimension (B=256, C=3, H=224, W=224) 100 times with a ViT backbone on one A100 gpu. This is simulated training time, so the rotation matrices were recalculated with every pass for learnable methods.

Note, Mixed RoPE is faster due to naive the use of naive vector partitioning operations and broadcasting. The main conclusion is that learning parameters and Spherical RoPE cause negligible computational overhead.

\subsection{Learned Frequencies}

When the frequencies of Spherical RoPE are learned, it is possible for the model to learn equivariance in a particular layer. Like Mixed RoPE, if the rotation frequencies in a layer is set to zero, then the attention score is position invariant. If one of the rotation frequencies is set to zero, then Spherical RoPE will become trivially equivariant in the remaining direction. This makes it interesting to observe what weights the model learns. Below we show the learned frequencies in each layer of the network after being trained on ImageNet-1k.

\begin{figure}
    \centering
    \includegraphics[width=\linewidth]{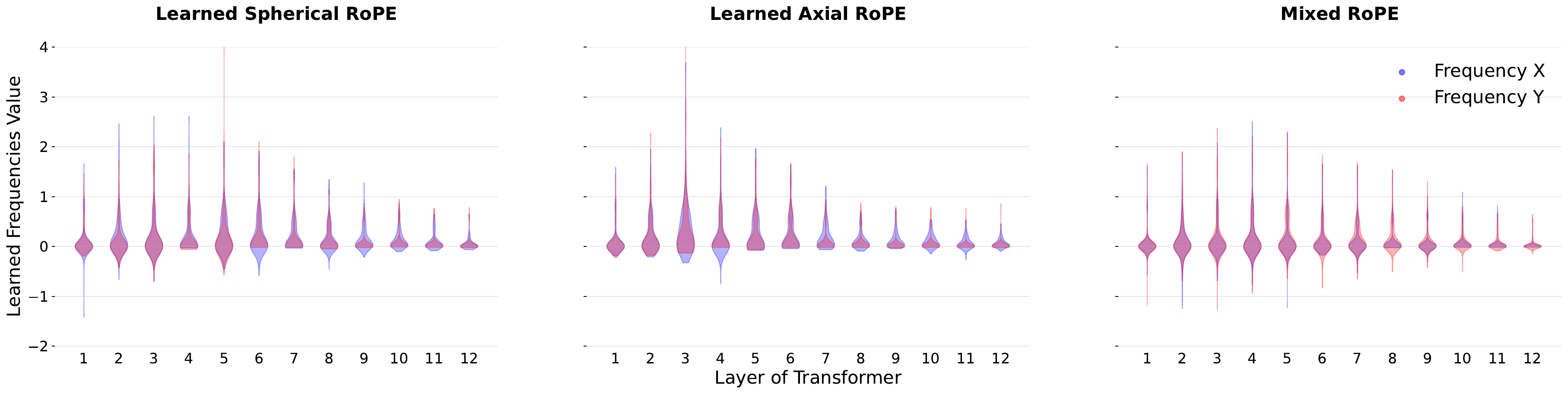}
    \caption{The distribution of learned frequencies in each layer of the ViT. Every method tends to learn low frequency positional encodings in the later layers of the network, meaning representations in the later layers are more invariant to position.}
    \label{fig:enter-label}
\end{figure}
\begin{figure}
    \centering
    \includegraphics[width=\linewidth]{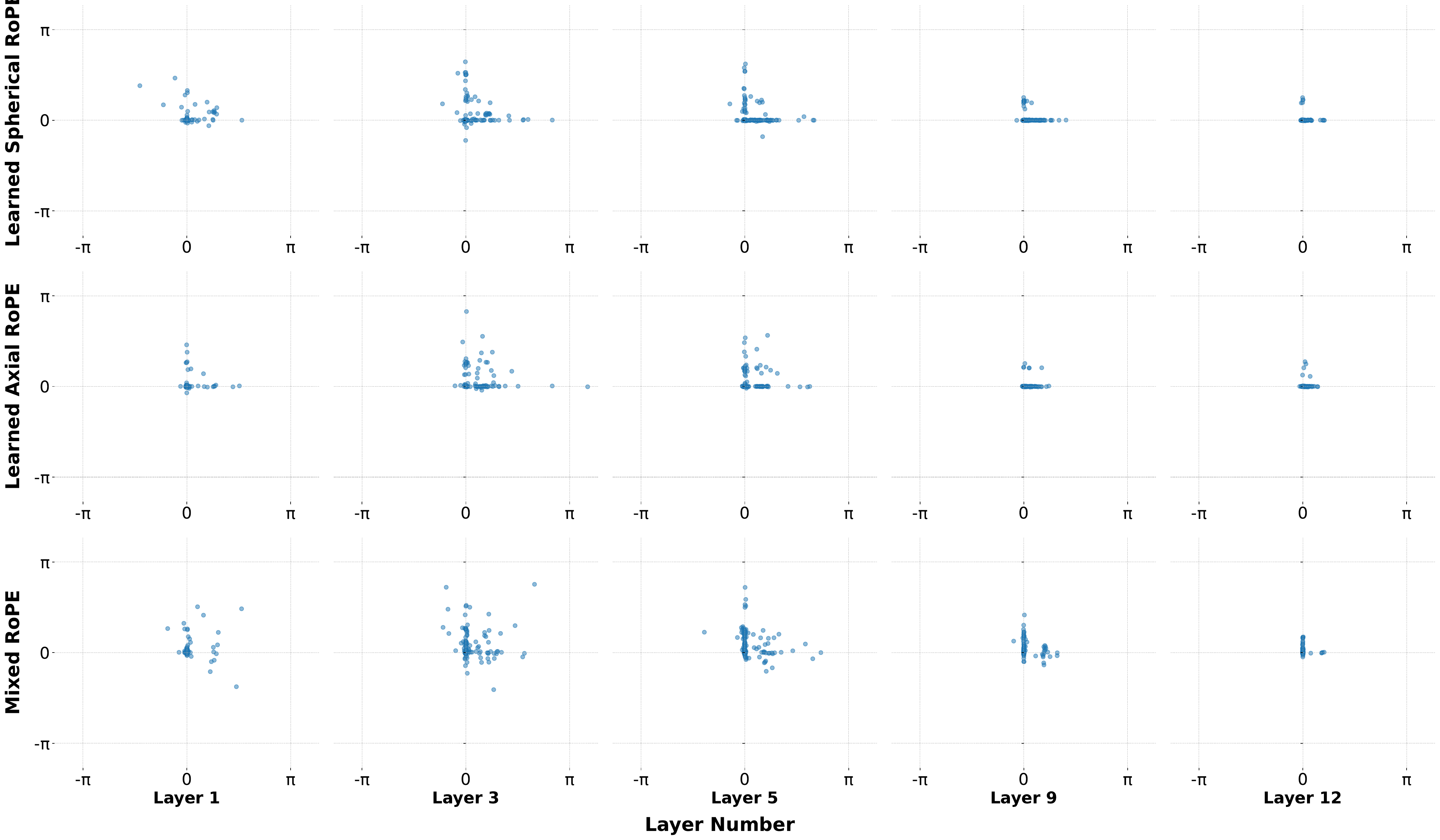}
    \caption{The scatterplot of learned $\omega_x$ and $\omega_y$. Note, though Axial RoPE is plotting $\omega_x$ and $\omega_y$ together, the rotations will always be axial, so there is no importance to the pairing.}
    \label{fig:enter-label}
\end{figure}

Because frequencies progressively trend toward the axis in deeper layers of the network which makes the positional encodings equivariant in that direction, one could argue that Spherical RoPE learns an equivariant representation in its later layers. However, this same trend can be seen in Mixed RoPE and more notably in Axial RoPE. Because Axial RoPE assumes mutual exclusivity, the frequency pairing is arbitrary. Since we still see the trend toward the axes, the observation that later layers use lower frequencies could be an artifact of backpropagation rather than a necessity for the model to learn an equivariant representation.

Interestingly, every method has notable clusters at zero frequencies. This suggests that much of the information in the images may be position agnostic. This further explains why setting low frequencies to zero in traditional RoPE improves performance as observed in \citet{barbero2024round}. An additional cluster can be observed most notably in the later layers of Spherical and Axial RoPE. We hypothesize this frequency corresponds to some information about the resolution of the image, i.\,e. the spacing of the grid. Some insight on how to generalize to higher resolutions may come from how this frequency corresponds to training data.

\newpage
\section{Proofs and Lemmas}

\paragraph{Axial RoPE Separability}
\vspace{30pt}
\begin{theorem}
    Axial RoPE is separable in $x$ and $y$, that is, the attention score can be decomposed into, 
    \begin{equation*}
        \alpha(\mathbf{x}_i, \mathbf{x}_j, \mathbf{p}_i, \mathbf{p}_j) = \alpha_{ij}^{(x)} + \alpha_{ij}^{(y)}
    \end{equation*}
\end{theorem}
\textbf{Proof.}
Suppose we define the dot‐product attention score as
\[
\alpha(\mathbf{q}, \mathbf{k}) = \mathbf{q}^\top \mathbf{k}.
\]
We incorporate \emph{Axial Rotary Positional Embeddings} by rotating each 2-dimensional subvector of the query (and likewise the key). Concretely, if the hidden dimension is \(2n\), we partition
\begin{equation}
\mathbf{q} = 
\bigl[\mathbf{q}_{x,1},\,\mathbf{q}_{y,1},\,\dots,\,\mathbf{q}_{x,n},\,\mathbf{q}_{y,n}\bigr]^\top,
\quad
\mathbf{k} = 
\bigl[\mathbf{k}_{x,1},\,\mathbf{k}_{y,1},\,\dots,\,\mathbf{k}_{x,n},\,\mathbf{k}_{y,n}\bigr]^\top,
\end{equation}
where each \(\mathbf{q}_{x,d},\,\mathbf{q}_{y,d},\,\mathbf{k}_{x,d},\,\mathbf{k}_{y,d}\in\mathbb{R}^2\). At spatial location \(\mathbf{p}=(p_x,p_y)\), we apply rotations
\[
\mathbf{q}'_{x,d} = \mathbf{R}\bigl(\omega_d\,p_x\bigr)\,\mathbf{q}_{x,d},
\quad
\mathbf{q}'_{y,d} = \mathbf{R}\bigl(\omega_d\,p_y\bigr)\,\mathbf{q}_{y,d},
\]
and similarly for \(\mathbf{k}\). Here \(\mathbf{R}(\theta)\in\mathbb{R}^{2\times2}\) is the planar rotation by angle \(\theta\).

For tokens at positions \(\mathbf{p}_i=(p_{i,x},p_{i,y})\) and \(\mathbf{p}_j=(p_{j,x},p_{j,y})\), their rotated queries and keys yield
\[
\alpha_{ij}
= \sum_{d=1}^{n}\Bigl[
  (\mathbf{q}_{x,d})^\top\,\mathbf{R}\bigl(\omega_d\,(p_{j,x}-p_{i,x})\bigr)\,\mathbf{k}_{x,d}
  +(\mathbf{q}_{y,d})^\top\,\mathbf{R}\bigl(\omega_d\,(p_{j,y}-p_{i,y})\bigr)\,\mathbf{k}_{y,d}
\Bigr].
\]
Define the horizontal and vertical components by
\[
\alpha_{ij}^{(x)}
:= \sum_{d=1}^{n} (\mathbf{q}_{x,d})^\top\,\mathbf{R}\bigl(\omega_d\,(p_{j,x}-p_{i,x})\bigr)\,\mathbf{k}_{x,d},
\quad
\alpha_{ij}^{(y)}
:= \sum_{d=1}^{n} (\mathbf{q}_{y,d})^\top\,\mathbf{R}\bigl(\omega_d\,(p_{j,y}-p_{i,y})\bigr)\,\mathbf{k}_{y,d}.
\]
Hence the total attention decomposes additively:
\[
\alpha_{ij} = \alpha_{ij}^{(x)} + \alpha_{ij}^{(y)},
\]
demonstrating that \emph{axial} rotary embeddings factorize the positional dependence along each axis.
\hfill\(\qed\)

\vspace{30pt}
\paragraph{Matrix Exponentiation}
Computing the matrix exponential by exponentiating the eigenvalues is a common result in linear algebra and numerics, however we provide it here for those unfamiliar.

\begin{lemma}
    Let $\mathbf{A}$ be a diagonalizable matrix $\mathbf{A} = \mathbf{U} \mathbf{\Lambda} \mathbf{U}^{-1}$, then the matrix exponential of $\mathbf{A}$ is given by
    \begin{equation*}
        \exp(\mathbf{A}) = \mathbf{U} \exp(\mathbf{\Lambda})~\mathbf{U}^{-1}
    \end{equation*} \label{lemma:exp}
\end{lemma}
\textbf{Proof.}

Recall the power‐series definition of the matrix exponential:
\begin{equation}
  \exp(\mathbf{A})
  = \sum_{k=0}^\infty \frac{1}{k!}\,\mathbf{A}^k.
\end{equation}
Since $\mathbf{A}$ is diagonalizable,
\begin{equation}
  \mathbf{A}^k
  = \bigl(\mathbf{U}\,\mathbf{\Lambda}\,\mathbf{U}^{-1}\bigr)^k
  = \mathbf{U}\,\mathbf{\Lambda}^k\,\mathbf{U}^{-1}.
\end{equation}
Substituting into the series gives
\begin{equation}
  \exp(\mathbf{A})
  = \sum_{k=0}^\infty \frac{1}{k!}\,
    \bigl(\mathbf{U}\,\mathbf{\Lambda}^k\,\mathbf{U}^{-1}\bigr)
  = \mathbf{U}
    \Bigl(\sum_{k=0}^\infty \frac{1}{k!}\,\mathbf{\Lambda}^k\Bigr)
    \mathbf{U}^{-1}.
\end{equation}
Because $\mathbf{\Lambda}$ is diagonal, the series
\(\sum_{k=0}^\infty \tfrac{1}{k!}\,\mathbf{\Lambda}^k\)
is itself the diagonal matrix of scalar exponentials,
\begin{equation}
  \exp(\mathbf{\Lambda})
  = \operatorname{diag}(e^{\lambda_1},\dots,e^{\lambda_n}).
\end{equation}
Hence is well defined, and
\begin{equation}
  \exp(\mathbf{A})
  = \mathbf{U}\,\exp(\mathbf{\Lambda})\,\mathbf{U}^{-1}.
\end{equation}
\hfill \qed

\paragraph{Simultaneous-Diagonalizability}
The proof that two (diagonalizable) matrixes are simultaneous-diagonalizability if and only if they are commutative is also a standard result. However, we once again provide it here:

\begin{lemma}\label{lemma:SimDiag}
Let $\mathcal{A}_x$ and $ \mathcal{A}_y$ be skew-symmetric. Then $\mathcal{A}_x$ and $ \mathcal{A}_y$ are simultaneously diagonalizable if and only if $\mathcal{A}_x\mathcal{A}_y = \mathcal{A}_y\mathcal{A}_x$ .
\end{lemma}
\textbf{Proof.}

Suppose $\mathcal{A}_x$ and $ \mathcal{A}_y$ are simultaneously diagonalizable. Then, because they are skew-symmetric, there exists a unitary matrix $\mathbf{U}$ such that
\begin{equation}
\mathbf{U} \Lambda_x \mathbf{U}^\top  = \mathcal{A}_x \quad \text{and} \quad \mathbf{U} \Lambda_y \mathbf{U}^\top  = \mathcal{A}_y,
\end{equation}
where $\Lambda_x$ and \( \Lambda_y \) are diagonal matrices.

Then,
\begin{equation}
    \mathcal{A}_x\mathcal{A}_y = \mathbf{U} \Lambda_x \mathbf{U}^\top \mathbf{U} \Lambda_y \mathbf{U}^\top = \mathbf{U} \Lambda_x \Lambda_y \mathbf{U}^\top = \mathbf{U} \Lambda_y \Lambda_x \mathbf{U}^\top = \mathcal{A}_x\mathcal{A}_y
\end{equation}

Hence, $\mathcal{A}_x$ and $\mathcal{A}_y$ commute.

Now suppose $\mathcal{A}_x$ and $\mathcal{A}_y$ commute,  $\mathcal{A}_x \mathcal{A}_y=\mathcal{A}_y \mathcal{A}_x$. Since  $\mathcal{A}_x$ and $\mathcal{A}_y$ are skew-symmetric, they are diagonalizable in $\mathbb{C}^{DxD}$, thus there exists a basis of eigenvectors of $\mathcal{A}_x$. Because $\mathcal{A}_y$ commutes with $\mathcal{A}_x$, the eigenspaces of $\mathcal{A}_x$ are invariant under $\mathcal{A}_y$. That is, for any eigenvalue $\lambda$ of $\mathcal{A}_x$, the corresponding eigenspace
\begin{equation}
E_\lambda = \{ v \in \mathbb{C}^D : \mathcal{A}_x v = \lambda v \}
\end{equation}
is $\mathcal{A}_y$-invariant: if $v \in E_\lambda $, then
\begin{equation}
\mathcal{A}_x(\mathcal{A}_y v) = \mathcal{A}_y(\mathcal{A}_x v) = \mathcal{A}_y(\lambda v) = \lambda \mathcal{A}_y v \Rightarrow \mathcal{A}_y v \in E_\lambda.
\end{equation}

Now, restrict $\mathcal{A}_x$ to each eigenspace $ E_\lambda $. Since $\mathbb{C}$ is algebraically closed and $ \mathcal{A}_y |_{E_\lambda} $ is a linear operator on a finite-dimensional space, $\mathcal{A}_y$ is diagonalizable on $E_\lambda$. Thus, we can choose a basis of eigenvectors for $\mathcal{A}_y$ in each  $E_\lambda$.

Putting these together, we get a basis for $\mathbb{C}^N$ consisting of vectors that are eigenvectors for both  $\mathcal{A}_x$ and $\mathcal{A}_y$. Therefore, $\mathcal{A}_x$ and $\mathcal{A}_y$ are simultaneously diagonalizable.

\hfill \qedsymbol

\paragraph{1-D LieRE is equivalent to RoPE}

In this section, we will more formally prove that the traditional RoPE with learned rotation frequencies is equivalent to 1-D RoPE as proposed in Section \ref{sec:generality}.

\begin{theorem_proof}
Any $D$-dimensional rotation can be parameterized by RoPE with learned frequencies.
\end{theorem_proof}

\textbf{Proof.}

We define a rotation to be an orthogonal matrix with positive determinant; that is, it is an element of $\mathbf{R}\in \text{SO}(N)$. We can write any element of $\text{SO}(N)$ via the exponential map $\mathbf{R} = e^\mathcal{A}$ where $\mathcal{A}\in \mathfrak{so}(N)$, i.e. $\mathcal{A}$ is a skew-symmetric matrix. It is well-known that the eigenvalues of a real, skew-symmetric matrix are purely imaginary (or zero), and such a matrix is unitarily (i.e. orthogonally) diagonalizable over $\mathbb{C}$, resulting in a spectral decomposition with a purely imaginary eigenvalue matrix. Thus,
\begin{equation}
    \mathcal{A} = \mathbf{U} \mathbf{\Lambda}i\mathbf{U}^\dagger
\end{equation}
and, by Lemma \ref{lemma:exp},
\begin{equation}
    \exp\left(\mathcal{A}\right) = \mathbf{U} \exp\left(\mathbf{\Lambda}i \right)\mathbf{U}^{\dagger}.
\end{equation}
where, because $\mathbf{\Lambda}$ is diagonal, $\exp(\mathbf{\Lambda})$ is simply the scalar-exponential of each element. The positional encoding of a token to a query can be written as,
\begin{equation}
    \varphi(\mathbf{x},p) = \exp(\mathcal{A}p) \mathbf{W}_q \mathbf{x} = \mathbf{U}\exp(\mathcal{\mathbf{\Lambda}}i~p)\mathbf{W_q' x} 
\end{equation}
where $\mathbf{W}'_q=\mathbf{W}_q\mathbf{U}$. We assume the same encoding for the key with a different matrix, $\mathbf{W}'_k$ and the same generator, $\mathcal{A}$. This equation can be rewritten as $\varphi(\mathbf{x},p)= \mathbf{U} RoPE(\mathbf{x},p)$ by Eq.\ref{eq:RoPE}. If the attention score is given by $\alpha(\mathbf{q},\mathbf{k}) = \mathbf{q}^{\dagger}\mathbf{k}$, where $\dagger$ denotes the Hermitian transpose, then the attention score can be expanded into,
\begin{align}
    \alpha(\mathbf{x}_i,\mathbf{x}_j,p_i,p_j) &= RoPE(\mathbf{x}_i,p_i)^\dagger\mathbf{U}^\dagger\mathbf{U}RoPE(\mathbf{x}_j,p_j) \\ &= RoPE(\mathbf{x}_i,p_i)^\dagger RoPE(\mathbf{x}_j,p_j).
\end{align}
Hence, \textit{any LieRE of one generator can be expressed as RoPE with learned rotation frequencies.} \qed

\paragraph{Any commutative LieRE is equivalent to Mixed RoPE}
We now prove that multi-dimensional LieRE with commutative generators generalizes directly to Mixed RoPE.

\begin{theorem_proof}
Any $M$-dimensional LieRE with commutative generators can be parameterized by Mixed RoPE.
\end{theorem_proof}

\textbf{Proof.}

Let ${\mathcal{A}_1, \dots, \mathcal{A}_M} \subset \mathfrak{so}(N)$ be skew-symmetric generators such that $[\mathcal{A}_m, \mathcal{A}_n] = \mathbf{0}$ for all $m,n$.
By Lemma~\ref{lemma:SimDiag}, commuting normal matrices are simultaneously unitarily diagonalizable.
Thus, there exists a unitary $\mathbf{U}$ and diagonal matrices $\mathbf{\Lambda}_1, \dots, \mathbf{\Lambda}_M$ such that
\begin{equation}
\mathcal{A}_m = \mathbf{U} \mathbf{\Lambda}_m i\mathbf{U}^\dagger
\quad\text{for all } m=1,\dots,M.
\end{equation}

For a position vector $\mathbf{p} = (p_1, \dots, p_M) \in \mathbb{R}^M$, the LieRE positional encoding is
\begin{align}
\mathrm{LieRE}(\mathbf{x},\mathbf{p})
&= \exp\left( \sum_{m=1}^M \mathcal{A}_m p_m \right) \mathbf{W}q \mathbf{x},
\end{align}
which, using Lemmas \ref{lemma:exp} and \ref{lemma:SimDiag}, can be written as
\begin{align}
    \mathrm{LieRE}(\mathbf{x},\mathbf{p})
&= \mathbf{U} \exp\left( \sum_{m=1}^M \mathbf{\Lambda}_m i,p_m \right) \mathbf{U}^\dagger \mathbf{W}_q \mathbf{x}.
\end{align}
Let $\mathbf{W}'_q = \mathbf{U}^\dagger \mathbf{W}_q$. Then
\begin{align}
\mathrm{LieRE}(\mathbf{x},\mathbf{p})
&= \mathbf{U} \mathrm{MixedRoPE}(\mathbf{x},\mathbf{p}),
\end{align}
where $\mathrm{MixedRoPE}$ applies elementwise complex rotations
\begin{align}
e^{i (\lambda_1^{(k)} p_1 + \dots + \lambda_M^{(k)} p_M)}
\end{align}
to each channel $k$, with frequencies ${\lambda_m^{(k)}}$ learned from $\mathbf{\Lambda}_m$.

If the attention score is given by $\alpha(\mathbf{q},\mathbf{k}) = \mathbf{q}^\dagger \mathbf{k}$, then
\begin{align}
\alpha(\mathbf{x}_i,\mathbf{x}_j,\mathbf{p}_i,\mathbf{p}_j)
&= \mathrm{MixedRoPE}(\mathbf{x}_i,\mathbf{p}_i)^\dagger \mathbf{U}^\dagger \mathbf{U} \mathrm{MixedRoPE}(\mathbf{x}_j,\mathbf{p}_j) \\
&= \mathrm{MixedRoPE}(\mathbf{x}_i,\mathbf{p}_i)^\dagger \mathrm{MixedRoPE}(\mathbf{x}_j\mathbf{p}_j).
\end{align}
Hence, \textit{any $M$-dimensional LieRE with commutative generators is equivalent to a Mixed RoPE parameterization with learned rotation frequencies.} \qed



\newpage
\section*{NeurIPS Paper Checklist}
\begin{enumerate}

\item {\bf Claims}
    \item[] Question: Do the main claims made in the abstract and introduction accurately reflect the paper's contributions and scope?
    \item[] Answer: \answerYes{} 
    \item[] Justification: We provide proofs and theoretical evidence on benchmarks.
    \item[] Guidelines:
    \begin{itemize}
        \item The answer NA means that the abstract and introduction do not include the claims made in the paper.
        \item The abstract and/or introduction should clearly state the claims made, including the contributions made in the paper and important assumptions and limitations. A No or NA answer to this question will not be perceived well by the reviewers. 
        \item The claims made should match theoretical and experimental results, and reflect how much the results can be expected to generalize to other settings. 
        \item It is fine to include aspirational goals as motivation as long as it is clear that these goals are not attained by the paper. 
    \end{itemize}

\item {\bf Limitations}
    \item[] Question: Does the paper discuss the limitations of the work performed by the authors?
    \item[] Answer: \answerYes{} 
    \item[] Justification: We emphasize that our conclusions are limited to vision and have a limitations section in the Appendix.
    \item[] Guidelines:
    \begin{itemize}
        \item The answer NA means that the paper has no limitation while the answer No means that the paper has limitations, but those are not discussed in the paper. 
        \item The authors are encouraged to create a separate "Limitations" section in their paper.
        \item The paper should point out any strong assumptions and how robust the results are to violations of these assumptions (e.g., independence assumptions, noiseless settings, model well-specification, asymptotic approximations only holding locally). The authors should reflect on how these assumptions might be violated in practice and what the implications would be.
        \item The authors should reflect on the scope of the claims made, e.g., if the approach was only tested on a few datasets or with a few runs. In general, empirical results often depend on implicit assumptions, which should be articulated.
        \item The authors should reflect on the factors that influence the performance of the approach. For example, a facial recognition algorithm may perform poorly when image resolution is low or images are taken in low lighting. Or a speech-to-text system might not be used reliably to provide closed captions for online lectures because it fails to handle technical jargon.
        \item The authors should discuss the computational efficiency of the proposed algorithms and how they scale with dataset size.
        \item If applicable, the authors should discuss possible limitations of their approach to address problems of privacy and fairness.
        \item While the authors might fear that complete honesty about limitations might be used by reviewers as grounds for rejection, a worse outcome might be that reviewers discover limitations that aren't acknowledged in the paper. The authors should use their best judgment and recognize that individual actions in favor of transparency play an important role in developing norms that preserve the integrity of the community. Reviewers will be specifically instructed to not penalize honesty concerning limitations.
    \end{itemize}

\item {\bf Theory assumptions and proofs}
    \item[] Question: For each theoretical result, does the paper provide the full set of assumptions and a complete (and correct) proof?
    \item[] Answer: \answerYes{} 
    \item[] Justification: While we do not formally list the assumptions, we implicitly make assumptions on positional encoding through assuming $N$-D LieRE.
    \item[] Guidelines:
    \begin{itemize}
        \item The answer NA means that the paper does not include theoretical results. 
        \item All the theorems, formulas, and proofs in the paper should be numbered and cross-referenced.
        \item All assumptions should be clearly stated or referenced in the statement of any theorems.
        \item The proofs can either appear in the main paper or the supplemental material, but if they appear in the supplemental material, the authors are encouraged to provide a short proof sketch to provide intuition. 
        \item Inversely, any informal proof provided in the core of the paper should be complemented by formal proofs provided in appendix or supplemental material.
        \item Theorems and Lemmas that the proof relies upon should be properly referenced. 
    \end{itemize}

    \item {\bf Experimental result reproducibility}
    \item[] Question: Does the paper fully disclose all the information needed to reproduce the main experimental results of the paper to the extent that it affects the main claims and/or conclusions of the paper (regardless of whether the code and data are provided or not)?
    \item[] Answer: \answerYes{} 
    \item[] Justification: We do our best to provide hyper-parameters for reproducing our results.
    \item[] Guidelines:
    \begin{itemize}
        \item The answer NA means that the paper does not include experiments.
        \item If the paper includes experiments, a No answer to this question will not be perceived well by the reviewers: Making the paper reproducible is important, regardless of whether the code and data are provided or not.
        \item If the contribution is a dataset and/or model, the authors should describe the steps taken to make their results reproducible or verifiable. 
        \item Depending on the contribution, reproducibility can be accomplished in various ways. For example, if the contribution is a novel architecture, describing the architecture fully might suffice, or if the contribution is a specific model and empirical evaluation, it may be necessary to either make it possible for others to replicate the model with the same dataset, or provide access to the model. In general. releasing code and data is often one good way to accomplish this, but reproducibility can also be provided via detailed instructions for how to replicate the results, access to a hosted model (e.g., in the case of a large language model), releasing of a model checkpoint, or other means that are appropriate to the research performed.
        \item While NeurIPS does not require releasing code, the conference does require all submissions to provide some reasonable avenue for reproducibility, which may depend on the nature of the contribution. For example
        \begin{enumerate}
            \item If the contribution is primarily a new algorithm, the paper should make it clear how to reproduce that algorithm.
            \item If the contribution is primarily a new model architecture, the paper should describe the architecture clearly and fully.
            \item If the contribution is a new model (e.g., a large language model), then there should either be a way to access this model for reproducing the results or a way to reproduce the model (e.g., with an open-source dataset or instructions for how to construct the dataset).
            \item We recognize that reproducibility may be tricky in some cases, in which case authors are welcome to describe the particular way they provide for reproducibility. In the case of closed-source models, it may be that access to the model is limited in some way (e.g., to registered users), but it should be possible for other researchers to have some path to reproducing or verifying the results.
        \end{enumerate}
    \end{itemize}

\item {\bf Open access to data and code}
    \item[] Question: Does the paper provide open access to the data and code, with sufficient instructions to faithfully reproduce the main experimental results, as described in supplemental material?
    \item[] Answer: \answerYes{} 
    \item[] Justification: 
    We intend to make the code public.
    \item[] Guidelines:
    \begin{itemize}
        \item The answer NA means that paper does not include experiments requiring code.
        \item Please see the NeurIPS code and data submission guidelines (\url{https://nips.cc/public/guides/CodeSubmissionPolicy}) for more details.
        \item While we encourage the release of code and data, we understand that this might not be possible, so “No” is an acceptable answer. Papers cannot be rejected simply for not including code, unless this is central to the contribution (e.g., for a new open-source benchmark).
        \item The instructions should contain the exact command and environment needed to run to reproduce the results. See the NeurIPS code and data submission guidelines (\url{https://nips.cc/public/guides/CodeSubmissionPolicy}) for more details.
        \item The authors should provide instructions on data access and preparation, including how to access the raw data, preprocessed data, intermediate data, and generated data, etc.
        \item The authors should provide scripts to reproduce all experimental results for the new proposed method and baselines. If only a subset of experiments are reproducible, they should state which ones are omitted from the script and why.
        \item At submission time, to preserve anonymity, the authors should release anonymized versions (if applicable).
        \item Providing as much information as possible in supplemental material (appended to the paper) is recommended, but including URLs to data and code is permitted.
    \end{itemize}

\item {\bf Experimental setting/details}
    \item[] Question: Does the paper specify all the training and test details (e.g., data splits, hyperparameters, how they were chosen, type of optimizer, etc.) necessary to understand the results?
    \item[] Answer: \answerYes{} 
    \item[] Justification: We provide them to the best of our ability.
    \item[] Guidelines:
    \begin{itemize}
        \item The answer NA means that the paper does not include experiments.
        \item The experimental setting should be presented in the core of the paper to a level of detail that is necessary to appreciate the results and make sense of them.
        \item The full details can be provided either with the code, in appendix, or as supplemental material.
    \end{itemize}

\item {\bf Experiment statistical significance}
    \item[] Question: Does the paper report error bars suitably and correctly defined or other appropriate information about the statistical significance of the experiments?
    \item[] Answer: \answerYes{} 
    \item[] Justification: We trained models from several random seeds.
    \item[] Guidelines:
    \begin{itemize}
        \item The answer NA means that the paper does not include experiments.
        \item The authors should answer "Yes" if the results are accompanied by error bars, confidence intervals, or statistical significance tests, at least for the experiments that support the main claims of the paper.
        \item The factors of variability that the error bars are capturing should be clearly stated (for example, train/test split, initialization, random drawing of some parameter, or overall run with given experimental conditions).
        \item The method for calculating the error bars should be explained (closed form formula, call to a library function, bootstrap, etc.)
        \item The assumptions made should be given (e.g., Normally distributed errors).
        \item It should be clear whether the error bar is the standard deviation or the standard error of the mean.
        \item It is OK to report 1-sigma error bars, but one should state it. The authors should preferably report a 2-sigma error bar than state that they have a 96\% CI, if the hypothesis of Normality of errors is not verified.
        \item For asymmetric distributions, the authors should be careful not to show in tables or figures symmetric error bars that would yield results that are out of range (e.g. negative error rates).
        \item If error bars are reported in tables or plots, The authors should explain in the text how they were calculated and reference the corresponding figures or tables in the text.
    \end{itemize}

\item {\bf Experiments compute resources}
    \item[] Question: For each experiment, does the paper provide sufficient information on the computer resources (type of compute workers, memory, time of execution) needed to reproduce the experiments?
    \item[] Answer: \answerYes{} 
    \item[] Justification: We provide basic information about the GPUs used.
    \item[] Guidelines:
    \begin{itemize}
        \item The answer NA means that the paper does not include experiments.
        \item The paper should indicate the type of compute workers CPU or GPU, internal cluster, or cloud provider, including relevant memory and storage.
        \item The paper should provide the amount of compute required for each of the individual experimental runs as well as estimate the total compute. 
        \item The paper should disclose whether the full research project required more compute than the experiments reported in the paper (e.g., preliminary or failed experiments that didn't make it into the paper). 
    \end{itemize}
    
\item {\bf Code of ethics}
    \item[] Question: Does the research conducted in the paper conform, in every respect, with the NeurIPS Code of Ethics \url{https://neurips.cc/public/EthicsGuidelines}?
    \item[] Answer: \answerYes{} 
    \item[] Justification: We do not believe there is any violations.
    \item[] Guidelines:
    \begin{itemize}
        \item The answer NA means that the authors have not reviewed the NeurIPS Code of Ethics.
        \item If the authors answer No, they should explain the special circumstances that require a deviation from the Code of Ethics.
        \item The authors should make sure to preserve anonymity (e.g., if there is a special consideration due to laws or regulations in their jurisdiction).
    \end{itemize}

\item {\bf Broader impacts}
    \item[] Question: Does the paper discuss both potential positive societal impacts and negative societal impacts of the work performed?
    \item[] Answer: \answerYes{} 
    \item[] Justification: We include a section in the appendix, however, it is mostly not applicable for our paper.
    \item[] Guidelines:
    \begin{itemize}
        \item The answer NA means that there is no societal impact of the work performed.
        \item If the authors answer NA or No, they should explain why their work has no societal impact or why the paper does not address societal impact.
        \item Examples of negative societal impacts include potential malicious or unintended uses (e.g., disinformation, generating fake profiles, surveillance), fairness considerations (e.g., deployment of technologies that could make decisions that unfairly impact specific groups), privacy considerations, and security considerations.
        \item The conference expects that many papers will be foundational research and not tied to particular applications, let alone deployments. However, if there is a direct path to any negative applications, the authors should point it out. For example, it is legitimate to point out that an improvement in the quality of generative models could be used to generate deepfakes for disinformation. On the other hand, it is not needed to point out that a generic algorithm for optimizing neural networks could enable people to train models that generate Deepfakes faster.
        \item The authors should consider possible harms that could arise when the technology is being used as intended and functioning correctly, harms that could arise when the technology is being used as intended but gives incorrect results, and harms following from (intentional or unintentional) misuse of the technology.
        \item If there are negative societal impacts, the authors could also discuss possible mitigation strategies (e.g., gated release of models, providing defenses in addition to attacks, mechanisms for monitoring misuse, mechanisms to monitor how a system learns from feedback over time, improving the efficiency and accessibility of ML).
    \end{itemize}
    
\item {\bf Safeguards}
    \item[] Question: Does the paper describe safeguards that have been put in place for responsible release of data or models that have a high risk for misuse (e.g., pretrained language models, image generators, or scraped datasets)?
    \item[] Answer: \answerNA{} 
    \item[] Justification: Our paper is more theoretical.
    \item[] Guidelines:
    \begin{itemize}
        \item The answer NA means that the paper poses no such risks.
        \item Released models that have a high risk for misuse or dual-use should be released with necessary safeguards to allow for controlled use of the model, for example by requiring that users adhere to usage guidelines or restrictions to access the model or implementing safety filters. 
        \item Datasets that have been scraped from the Internet could pose safety risks. The authors should describe how they avoided releasing unsafe images.
        \item We recognize that providing effective safeguards is challenging, and many papers do not require this, but we encourage authors to take this into account and make a best faith effort.
    \end{itemize}

\item {\bf Licenses for existing assets}
    \item[] Question: Are the creators or original owners of assets (e.g., code, data, models), used in the paper, properly credited and are the license and terms of use explicitly mentioned and properly respected?
    \item[] Answer: \answerNA{} 
    \item[] Justification: We cite libraries used and datasets, however they are standard libraries and benchmarks. There are no other specialized assets used.
    \item[] Guidelines:
    \begin{itemize}
        \item The answer NA means that the paper does not use existing assets.
        \item The authors should cite the original paper that produced the code package or dataset.
        \item The authors should state which version of the asset is used and, if possible, include a URL.
        \item The name of the license (e.g., CC-BY 4.0) should be included for each asset.
        \item For scraped data from a particular source (e.g., website), the copyright and terms of service of that source should be provided.
        \item If assets are released, the license, copyright information, and terms of use in the package should be provided. For popular datasets, \url{paperswithcode.com/datasets} has curated licenses for some datasets. Their licensing guide can help determine the license of a dataset.
        \item For existing datasets that are re-packaged, both the original license and the license of the derived asset (if it has changed) should be provided.
        \item If this information is not available online, the authors are encouraged to reach out to the asset's creators.
    \end{itemize}

\item {\bf New assets}
    \item[] Question: Are new assets introduced in the paper well documented and is the documentation provided alongside the assets?
    \item[] Answer: \answerNA{} 
    \item[] Justification: 
    \item[] Guidelines:
    \begin{itemize}
        \item The answer NA means that the paper does not release new assets.
        \item Researchers should communicate the details of the dataset/code/model as part of their submissions via structured templates. This includes details about training, license, limitations, etc. 
        \item The paper should discuss whether and how consent was obtained from people whose asset is used.
        \item At submission time, remember to anonymize your assets (if applicable). You can either create an anonymized URL or include an anonymized zip file.
    \end{itemize}

\item {\bf Crowdsourcing and research with human subjects}
    \item[] Question: For crowdsourcing experiments and research with human subjects, does the paper include the full text of instructions given to participants and screenshots, if applicable, as well as details about compensation (if any)? 
    \item[] Answer: \answerNA{} 
    \item[] Justification:
    \item[] Guidelines:
    \begin{itemize}
        \item The answer NA means that the paper does not involve crowdsourcing nor research with human subjects.
        \item Including this information in the supplemental material is fine, but if the main contribution of the paper involves human subjects, then as much detail as possible should be included in the main paper. 
        \item According to the NeurIPS Code of Ethics, workers involved in data collection, curation, or other labor should be paid at least the minimum wage in the country of the data collector. 
    \end{itemize}

\item {\bf Institutional review board (IRB) approvals or equivalent for research with human subjects}
    \item[] Question: Does the paper describe potential risks incurred by study participants, whether such risks were disclosed to the subjects, and whether Institutional Review Board (IRB) approvals (or an equivalent approval/review based on the requirements of your country or institution) were obtained?
    \item[] Answer: \answerNA{} 
    \item[] Justification:
    \item[] Guidelines:
    \begin{itemize}
        \item The answer NA means that the paper does not involve crowdsourcing nor research with human subjects.
        \item Depending on the country in which research is conducted, IRB approval (or equivalent) may be required for any human subjects research. If you obtained IRB approval, you should clearly state this in the paper. 
        \item We recognize that the procedures for this may vary significantly between institutions and locations, and we expect authors to adhere to the NeurIPS Code of Ethics and the guidelines for their institution. 
        \item For initial submissions, do not include any information that would break anonymity (if applicable), such as the institution conducting the review.
    \end{itemize}

\item {\bf Declaration of LLM usage}
    \item[] Question: Does the paper describe the usage of LLMs if it is an important, original, or non-standard component of the core methods in this research? Note that if the LLM is used only for writing, editing, or formatting purposes and does not impact the core methodology, scientific rigorousness, or originality of the research, declaration is not required.
    \item[] Answer: \answerNA{} 
    \item[] Justification:
    \item[] Guidelines:
    \begin{itemize}
        \item The answer NA means that the core method development in this research does not involve LLMs as any important, original, or non-standard components.
        \item Please refer to our LLM policy (\url{https://neurips.cc/Conferences/2025/LLM}) for what should or should not be described.
    \end{itemize}

\end{enumerate}

\end{document}